\documentclass[acmtog]{acmart}

\AtBeginDocument{%
  \providecommand\BibTeX{{%
    \normalfont B\kern-0.5em{\scshape i\kern-0.25em b}\kern-0.8em\TeX}}}

\setcopyright{none}
\acmJournal{TOG}
\acmYear{2024}
\acmVolume{43}
\acmNumber{6}
\acmArticle{}
\acmMonth{12}
\acmDOI{10.1145/3687759}

\usepackage{graphicx}
\usepackage{amsmath} 
\usepackage{color}
\usepackage{colortbl}
\usepackage{xcolor}
\usepackage{multirow}
\usepackage{soul}
\usepackage{pifont}
\usepackage{hyperref}
\usepackage{enumitem}
\usepackage{algorithm}
\usepackage{algorithmic}
\makeatletter
\DeclareRobustCommand\onedot{\futurelet\@let@token\@onedot}
\def\@onedot{\ifx\@let@token.\else.\null\fi\xspace}

\def\eg{\emph{e.g}\onedot} 
\def\ie{\emph{i.e}\onedot}

\def\deg{$^\circ$~}
\def\gs{3D Gaussians~}
\makeatother

\definecolor{metric-1}{HTML}{ff9999}
\definecolor{metric-2}{HTML}{ffcc99}
\definecolor{metric-3}{HTML}{fff6b2}

\usepackage{xspace}
\def\name{GaussianObject\@\xspace}
\usepackage{soul}
\newcommand{\rev}[1]{#1}
\begin{document}

\title{GaussianObject: High-Quality 3D Object Reconstruction from Four Views with Gaussian Splatting}

\author{Chen Yang}
\authornote{Equal contributions.}
\orcid{0000-0003-4496-7849}
\affiliation{%
  \institution{MoE Key Lab of Artificial Intelligence, AI Institute, SJTU}
  \city{Shanghai}
  \country{China}
  }
\email{ycyangchen@sjtu.edu.cn}

\author{Sikuang Li}
\authornotemark[1]
\orcid{0009-0008-4080-7454}
\affiliation{%
  \institution{MoE Key Lab of Artificial Intelligence, AI Institute, SJTU}
  \city{Shanghai}
   \country{China}
  }
\email{uranusits@sjtu.edu.cn}

\author{Jiemin Fang}
\authornote{Project lead.}
\orcid{0000-0002-0322-4582}
\affiliation{%
  \institution{Huawei Inc.}
  \city{Wuhan}
  \country{China}
  }
\email{jaminfong@gmail.com}

\author{Ruofan Liang}
\orcid{0009-0005-7667-1809}
\affiliation{%
  \institution{University of Toronto}
  \city{Toronto}
  \country{Canada}
  }
\email{ruofan@cs.toronto.edu}

\author{Lingxi Xie}
\orcid{0000-0003-4831-9451}
\affiliation{%
  \institution{Huawei Inc.}
  \city{Beijing}
   \country{China}
  }
\email{198808xc@gmail.com}

\author{Xiaopeng Zhang}
\orcid{0000-0001-6337-5748}
\affiliation{%
  \institution{Huawei Inc.}
  \city{Shanghai}
   \country{China}
  }
\email{zxphistory@gmail.com}

\author{Wei Shen}
\authornote{Corresponding author.}
\orcid{0000-0002-1235-598X}
\affiliation{%
  \institution{MoE Key Lab of Artificial Intelligence, AI Institute, SJTU}
  \city{Shanghai}
   \country{China}
  }
\email{wei.shen@sjtu.edu.cn}

\author{Qi Tian}
\orcid{0000-0002-7252-5047}
\affiliation{%
  \institution{Huawei Inc.}
  \city{Shenzhen}
   \country{China}
  }
\email{tian.qi1@huawei.com}

\renewcommand\shortauthors{Yang, Li, et al.}

\begin{abstract}
Reconstructing and rendering 3D objects from highly sparse views is of critical importance for promoting applications of 3D vision techniques and improving user experience. However, images from sparse views only contain very limited 3D information, leading to two significant challenges: 1) Difficulty in building multi-view consistency as images for matching are too few; 2) Partially omitted or highly compressed object information as view coverage is insufficient. To tackle these challenges, we propose \name, a framework to represent and render the 3D object with Gaussian splatting that achieves high rendering quality with only 4 input images. We first introduce techniques of visual hull and floater elimination, which explicitly inject structure priors into the initial optimization process to help build multi-view consistency, yielding a coarse 3D Gaussian representation. Then we construct a Gaussian repair model based on diffusion models to supplement the omitted object information, where Gaussians are further refined. We design a self-generating strategy to obtain image pairs for training the repair model. We further design a COLMAP-free variant, where pre-given accurate camera poses are not required, which achieves competitive quality and facilitates wider applications. \name is evaluated on several challenging datasets, including MipNeRF360, OmniObject3D, OpenIllumination, and our-collected unposed images, achieving superior performance from only four views and significantly outperforming previous SOTA methods. Our demo is available at \textcolor{blue}{\textit{\url{https://gaussianobject.github.io/}}}, and the code has been released at \textcolor{blue}{\textit{\url{https://github.com/GaussianObject/GaussianObject}}}.
\end{abstract}



\begin{teaserfigure}
  \includegraphics[width=\textwidth]{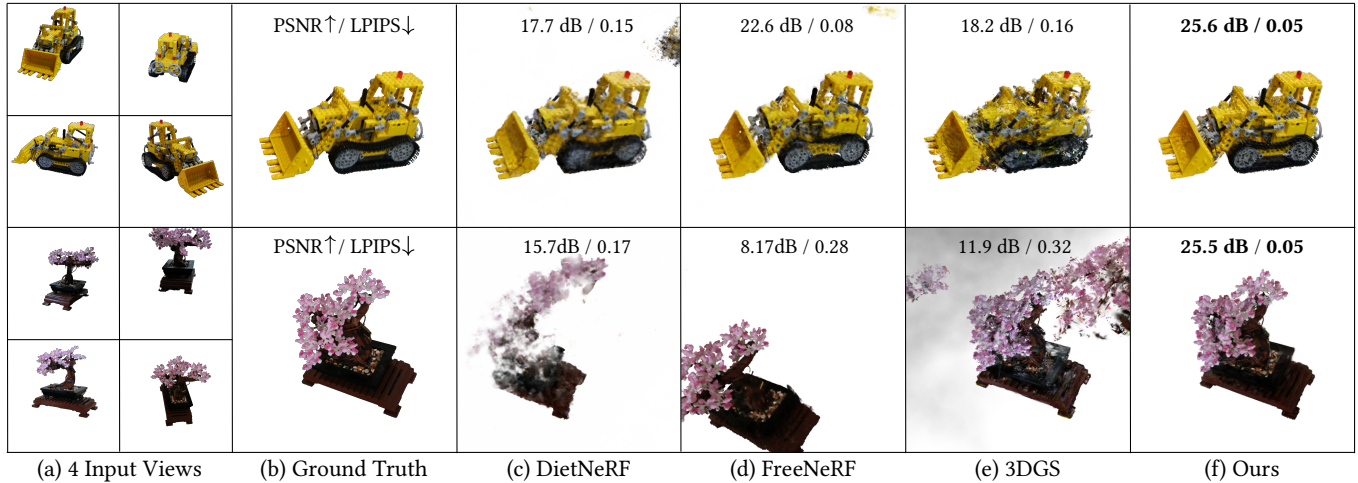}
  \vspace{-1em}
  \caption{We introduce GaussianObject, a framework capable of reconstructing high-quality 3D objects from only 4 images with Gaussian splatting. GaussianObject demonstrates superior performance over previous state-of-the-art (SOTA) methods on challenging objects.}
  \Description{We introduce GaussianObject, a framework capable of reconstructing high-quality 3D objects from only 4 images with Gaussian splatting. GaussianObject demonstrates superior performance over previous SOTA methods on challenging objects.}
  \label{fig:teaser}
\end{teaserfigure}

\maketitle

\section{Introduction}
Reconstructing and rendering 3D objects from 2D images has been a long-standing and important topic, which plays critical roles in a vast range of real-life applications. One key factor that impedes users, especially ones without expert knowledge, from widely using these techniques is that usually dozens of multi-view images need to be captured, which is cumbersome and sometimes impractical. Efficiently reconstructing high-quality 3D objects from highly sparse captured images is of great value for expediting downstream applications such as 3D asset creation for game/movie production and AR/VR products.

In recent years, a series of methods~\cite{dietnerf, regnerf, sparsenerf, freenerf, zerorf, fsgs, darf, sparsefusion} have been proposed to reduce reliance on dense captures. However, it is still challenging to produce high-quality 3D objects when the views become \textbf{extremely sparse}, \eg only 4 images in a 360\deg range, as shown in Fig.~\ref{fig:teaser}. We delve into the task of sparse-view reconstruction and discover two main challenges behind it. The first one lies in the difficulty of building multi-view consistency from highly sparse input. The 3D representation is easy to overfit the input images and degrades into fragmented pixel patches of training views without reasonable structures.
The other challenge is that with sparse captures in a 360\deg range, some content of the object can be inevitably omitted or severely compressed when observed from extreme views\footnote{When the view is orthogonal to the surface of the object, the observed information attached to the surface can be largely \rev{preserved}; On the contrary, the information will be severely compressed.}. The omitted or compressed information is impossible or hard to be reconstructed in 3D only from the input images. 

To tackle the aforementioned challenges, we introduce \name, a novel framework designed to reconstruct high-quality 3D objects from as few as 4 input images. 
We choose 3D Gaussian splatting (3DGS)~\cite{3dgs} as the basic representation as it is fast and, more importantly, explicit enough. Benefiting from its point-like structure, we design several techniques for introducing object structure priors, \eg the basic/rough geometry of the object, to help build multi-view consistency, including visual hull~\cite{visual_hull} to locate Gaussians within the object outline and floater elimination to remove outliers. 
To erase artifacts caused by omitted or highly compressed object information, we propose a Gaussian repair model driven by 2D large diffusion models~\cite{ldm}, translating corrupted rendered images into high-fidelity ones. As normal diffusion models lack the ability to repair corrupted images, we design self-generating strategies to construct image pairs to tune the diffusion models, including rendering images from leave-one-out training models and adding 3D noises to Gaussian attributes. Images generated from the repair model can be used to refine the 3D Gaussians optimized with structure priors, where the rendering quality can be further improved. 
To further extend GaussianObject to practical applications, we introduce a COLMAP-free variant of GaussianObject (CF-GaussianObject), which achieves competitive reconstruction performance on challenging datasets with only four input images without inputting accurate camera parameters.

Our contributions are summarized as follows:
\begin{itemize}[leftmargin=12pt]
\item We optimize 3D Gaussians from highly sparse views using explicit structure priors, introducing techniques of visual hull for initialization and floater elimination for training.
\item We propose a Gaussian repair model based on diffusion models to remove artifacts caused by omitted or highly compressed information, where the rendering quality can be further improved.
\item The overall framework \name consistently outperforms current SOTA methods on several challenging real-world datasets, both qualitatively and quantitatively. A COLMAP-free variant is further presented for wider applications, weakening the requirement of accurate camera poses.
\end{itemize}

\begin{figure*}[thbp]
    \centering
    \includegraphics[width=1.0\linewidth]{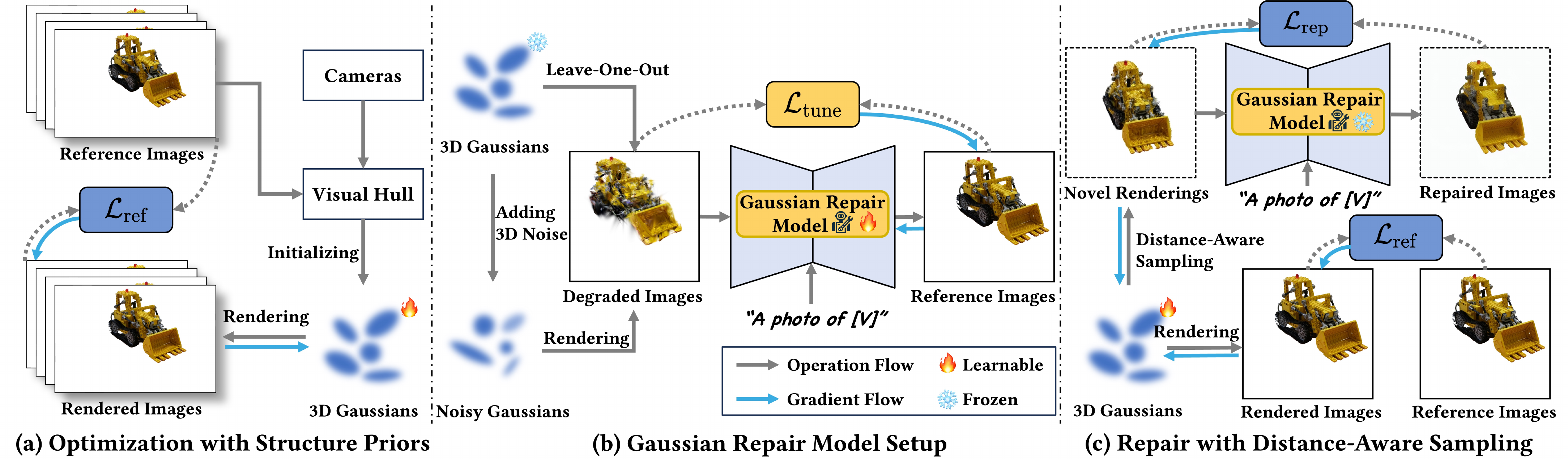}
    \vspace{-2em}
    \caption{Overview of GaussianObject.
    (a) We initialize 3D Gaussians by constructing a visual hull with camera parameters and masked images, which are optimized with $\mathcal{L}_{\text{ref}}$ and refined through floater elimination. (b) We use a novel `leave-one-out' strategy and add 3D noise to Gaussians to generate corrupted Gaussian renderings. These renderings, paired with their corresponding reference images, facilitate the training of the Gaussian repair model employing $\mathcal{L}_{\text{tune}}$. For details please refer to Fig. \ref{fig: repair_model illustration}.
    (c) Once trained, the Gaussian repair model is frozen and used to correct views that need to be rectified. These views are identified through distance-aware sampling. The repaired images and reference images are used to further optimize 3D Gaussians with $\mathcal{L}_{\text{rep}}$ and $\mathcal{L}_{\text{ref}}$.}
    \Description{}
    \label{fig: framework}
    \vspace{-0.5em}
\end{figure*}

\section{Related Work}

Vanilla NeRF struggles in sparse settings. Techniques like \citet{dsnerf, densedepthprior, vipnerf, simplenerf, somraj2024simple} use Structure from Motion (SfM)~\cite{sfm} derived visibility or depth \rev{and} mainly focus on closely aligned views. \citet{sinnerf} uses ground truth depth maps, which are costly to obtain in real-world images. Some methods~\cite{darf, sparsenerf} estimate depths with monocular depth estimation models~\cite{midas, dpt} or sensors, but these are often too coarse. \citet{dietnerf} uses a vision-language model~\cite{clip} for unseen view rendering, but the semantic consistency is too high-level to guide low-level reconstruction. 
\citet{zerorf} combines a deep image prior with factorized NeRF, effectively capturing overall appearance but missing fine details in input views.
Priors based on information theory~\cite{infonerf}, continuity~\cite{regnerf}, symmetry~\cite{flipnerf}, and frequency regularization~\cite{freenerf, halo} are \rev{only} effective for specific scenarios, limiting their further applications. Besides, there are some methods~\cite{leap, nvist, agg, tgs} that employ Vision Transformer (ViT)~\cite{vit} to reduce the requirements for constructing NeRFs and Gaussians.

The recent progress in diffusion models has spurred notable advancements in 3D applications. Dreamfusion~\cite{dreamfusion} proposes Score Distillation Sampling (SDS) for distilling NeRFs with 2D priors from a pre-trained diffusion model for 3D object generation from text prompts. It has been further refined for text-to-3D~\cite{magic3d, latentnerf, sjc, fantasia3d, prolificdreamer, gaussiandreamer, dreamgaussian, mvdream} and 3D/4D editing~\cite{instructn2n, control4d} by various studies, demonstrating the versatility of 2D diffusion models in 3D contexts. \citet{hifa, genvs, zero123, viewneti, efficient4d, muller2024multidiff} have adapted these methods for 3D generation and view synthesis from a single image, while they often have strict input requirements and can produce overly saturated images. 
In sparse reconstruction, approaches like DiffusioNeRF~\cite{diffusionerf}, SparseFusion~\cite{sparsefusion}, Deceptive-NeRF~\cite{deceptivenerf}, ReconFusion~\cite{reconfusion} and CAT3D~\cite{gao2024cat3d} integrate diffusion models with NeRFs. 
Recently, Large reconstruction models (LRMs)~\cite{lrm,pf-lrm,dmv3d,meshlrm,grm,Instant3d_fast,lgm,humanlrm2023,gslrm2024} also achieve 3D reconstruction from highly sparse views. 
\rev{Though effective in generating images fast, these methods encounter issues with large pretraining, strict requirements on view distribution and object location, and difficulty in handling real-world captures.}

While 3DGS shows strong power in novel view synthesis, it struggles with sparse 360\deg views similar to NeRF. Inspired by few-shot NeRFs, methods~\cite{fsgs, depth_reg_gs, sparsegs, paliwal2024coherentgs, charatan2024pixelsplat} have been developed for sparse 360\deg reconstruction. However, they still severely rely on the SfM points. 
Our \name proposes structure-prior-aided Gaussian initialization to tackle this issue, drastically reducing the required input views to only 4, a significant improvement compared with over 20 views required by FSGS~\cite{fsgs}.

\section{Method}
The subsequent sections detail the methodology: Sec.~\ref{Method: Preliminary} reviews foundational techniques; Sec.~\ref{Method: framework} introduces our overall framework;
Sec.~\ref{Method: Initial Optimization with Structure Priors} describes how we apply the structure priors for initial optimization; Sec.~\ref{Method: Gaussian Repair Model Setup} details the setup of our Gaussian repair model; Sec.~\ref{Method: Gaussian Repair with Distance-Aware Sampling} illustrates the repair of \gs using this model and Sec.~\ref{Method: COLMAP-free  GaussianObject} elucidates the COLMAP-free version of \name. 
\rev{To facilitate a better understanding, all key mathematical symbols and their corresponding meanings are listed in Table~\ref{tab:math-symbols}.}

\begin{table}[tbp]
\centering
\caption{\rev{List of Key Mathematical Symbols}}
\label{tab:math-symbols}
\begin{tabular}{ll}
\hline
Symbol & Meaning \\
\hline
\( X^{\text{ref}} = \{ x_i \}_{i=1}^N \)            & Reference images \\
\( K^{\text{ref}} = \{ k_i \}_{i=1}^N \)            & Intrinsics of \( X^{\text{ref}}\) \\
\(\hat{K}^{\text{ref}}  \)                          & Estimated intrinsics of \( X^{\text{ref}}\) \\
\(\hat{K}\)                                         & Estimated shared intrinsics of \( X^{\text{ref}}\) \\
\( \Pi^{\text{ref}} = \{ \pi_i \}_{i=1}^N \)        & Extrinsics of \( X^{\text{ref}}\) \\
\( \Pi^{\text{nov}}  \)                             & Extrinsics of viewpoints in repair path \\
\( \hat{\Pi}^{\text{ref}} \)                        & Estimated extrinsics of \( X^{\text{ref}}\) \\
\( M^{\text{ref}} = \{ m_i \}_{i=1}^N \)            & Masks of \( X^{\text{ref}}\) \\
\(\mu\)                                            & Center location of Gaussian \\
\(q\)                                               & Rotation quaternion of Gaussian \\
\(s\)                                              & Scale vector of Gaussian \\
\(\sigma\)                                         & Opacity of Gaussian \\
\(sh\)                                             & Spherical harmonic coefficients of Gaussian \\
\(\mathcal{G}_c\)                                  & Coarse 3D Gaussians \\
\(\mathcal{R}\)                                    & Diffusion based Gaussian repair model \\
\(\mathcal{E}\)                                     & Latent diffusion encoder of \(\mathcal{R}\) \\
\(\mathcal{D}\)                                     & Latent diffusion decoder of \(\mathcal{R}\) \\
\(x^\prime\)                                       & Degraded rendering \\
\(\hat{x}\)                                        & Image repaired by \(\mathcal{R}\) \\
\(\epsilon_s\)                                      & 3D Noise added to attributes of \(\mathcal{G}_c\) \\
\(\epsilon\)                                        & 2D Gaussian noise for fine-tuning \\
\(\epsilon_\theta\)                                 & 2D Noise predicted by \(\mathcal{R}\) \\
\(c^{\text{tex}}\)                                  & Object-specific language prompt \\
\(\mathcal{P}\)                                     & Coarse point cloud predicted by DUSt3R \\
\hline
\end{tabular}
\end{table}

\subsection{Preliminary} \label{Method: Preliminary}
\textit{3D Gaussian Splatting.}
3D Gaussian Splatting~\cite{3dgs} represents 3D scenes with 3D Gaussians. Each 3D Gaussian is composed of the center location $\mu$, rotation quaternion $q$, scaling vector $s$, opacity $\sigma$, and spherical harmonic (SH) coefficients $sh$. 
Thus, a scene is parameterized as a set of Gaussians $\mathcal{G}=\{G_i:\mu_i,q_i,s_i,\sigma_i,sh_i\}_{i=1}^{P}$. 

\textit{ControlNet.}
Diffusion models are generative models that sample from a data distribution $q(X_0)$, beginning with Gaussian noise $\epsilon$ and using various sampling schedulers. They operate by reversing a discrete-time stochastic noise addition process $\{X_t\}_{t=0}^T$ with a diffusion model $p_\theta(X_{t-1}|X_t)$ trained to approximate $q(X_{t-1}|X_t)$, where $t \in [0, T]$ is the noise level and $\theta$ is the learnable parameters.
Substituting $X_0$ with its latent code $Z_0$ from a Variational Autoencoder (VAE)~\cite{vae} leads to the development of Latent Diffusion Models (LDM)~\cite{ldm}. ControlNet~\cite{controlnet} further enhances the generative process with additional image conditioning by integrating a network structure similar to the diffusion model, optimized with the loss function: 
\begin{equation} \label{eq: controlnet loss}
    \mathcal{L}_{Cond}=\mathbb{E}_{Z_0,t,\epsilon}[\|\epsilon_\theta(\sqrt{\bar{\alpha}_t}Z_0+\sqrt{1-\bar{\alpha}_t}\epsilon,t,c^{\text{tex}},c^{\text{img}})-\epsilon\|_2^2],
\end{equation}
where $c^{\text{tex}}$ and $c^{\text{img}}$ denote the text and image conditioning respectively, and $\epsilon_\theta$ is the Gaussian noise inferred by the diffusion model with parameter $\theta$, $\bar{\alpha}_{1:T}\in(0,1]^T$ is a decreasing sequence associated with the noise-adding process.

\subsection{Overall Framework} \label{Method: framework}
Given a sparse collection of $N$ reference images \( X^{\text{ref}} = \{ x_i \}_{i=1}^N \), captured within a 360\deg range and encompassing one object, along with the corresponding camera \rev{intrinsics\footnote{\rev{Given that the camera intrinsics are known and fixed, we exclude them from the rendering function for simplicity.}} \( K^{\text{ref}} = \{ k_i \}_{i=1}^N \), extrinsics \( \Pi^{\text{ref}} = \{ \pi_i \}_{i=1}^N \)} and masks \( M^{\text{ref}} = \{ m_i \}_{i=1}^N \) of the object, our target is to obtain a 3D representation $\mathcal{G}$, which can achieve photo-realistic rendering \( x = \mathcal{G}(\pi | \{ x_i, \pi_i, m_i \}_{i=1}^N) \) from any viewpoint.
To achieve this, we employ the 3DGS model for its simplicity for structure priors embedding and fast rendering capabilities. 
The process begins with initializing 3D Gaussians using a visual hull~\cite{visual_hull}, followed by optimization with floater elimination, enhancing the structure of Gaussians.
Then we design self-generating strategies to supply sufficient image pairs for constructing a Gaussian repair model, which is used to rectify incomplete object information. The overall framework is shown in Fig.~\ref{fig: framework}.

\subsection{Initial Optimization with Structure Priors}
\label{Method: Initial Optimization with Structure Priors}
Sparse views, especially for only 4 images, provide limited 3D information for reconstruction. 
In this case, SfM points, which are the key for 3DGS initialization, are often absent. 
Besides, insufficient multi-view consistency leads to ambiguity among shape and appearance, resulting in many floaters during reconstruction.
We propose two techniques to initially optimize the 3D Gaussian representation, which take full advantage of structure priors from the limited views and result in a satisfactory outline of the object.

\textit{Initialization with Visual Hull.}
To better leverage object structure information from limited reference images, we utilize the view frustums and object masks to create a visual hull as a geometric scaffold for initializing our 3D Gaussians. 
Compared with \rev{the limited number of SfM points in extremely sparse settings,} the visual hull provides more structure priors that help build multiview consistency by excluding unreasonable Gaussian distributions.
The cost of the visual hull is just several masks derived from sparse 360\deg images, which can be easily acquired using current segmentation models such as SAM~\cite{sam}.
Specifically, points are randomly initialized within the visual hull using rejection sampling: we project uniformly sampled random 3D points onto image planes and retain those within the intersection of all image-space masks.
Point colors are averaged from bilinearly interpolated pixel colors across reference image projections.
Then we transform these 3D points into 3D Gaussians.
For each point, we assign its position as $\mu$ and convert its color into $sh$. The mean distance between adjacent points forms the scale $s$, while the rotation $q$ is set to a unit quaternion as default. 
The opacity $\sigma$ is initialized to a constant value.
This initialization strategy relies on the initial masks. Despite potential inaccuracies in these masks or unrepresented concavities by the visual hull, we observed that subsequent optimization processes reliably yield high-quality reconstructions.

\textit{Floater Elimination.}
While the visual hull builds a coarse estimation of the object geometry, it often contains regions that do not belong to the object due to the inadequate coverage of reference images. These regions usually appear to be floaters, damaging the quality of novel view synthesis.
These floaters are problematic as the optimization process struggles to adjust them due to insufficient observational data regarding their position and appearance.

To mitigate this issue, we utilize the statistical distribution of distances among the 3D Gaussians to distinguish the primary object and the floaters. This is implemented by the K-Nearest Neighbors (KNN) algorithm, which calculates the average distance to the nearest $\sqrt{P}$ Gaussians for each element in $\mathcal{G}_c$. We then establish a normative range by computing the mean and standard deviation of these distances.
\rev{Based on statistical analysis, we exclude Gaussians whose mean neighbor distances exceed the adaptive threshold $\tau = \text{mean} + \lambda_e \text{std}$. This thresholding process is repeated periodically throughout optimization, where $\lambda_e$ is linearly decreased to 0 to refine the scene representation progressively.}

\textit{Initial Optimization}
The optimization of $\mathcal{G}_c$ incorporates color, mask, and monocular depth losses. The color loss combines L1 and D-SSIM losses from 3D Gaussian Splatting:
\begin{equation}
\mathcal{L}_{1} = \|x - x^{\text{ref}}\|_1, \quad \mathcal{L}_{\text{D-SSIM}} = 1 - \text{SSIM}(x, x^{\text{ref}}),
\end{equation}
where $x$ is the rendering and $x^{\text{ref}}$ is the corresponding reference image.
A binary cross entropy (BCE) loss~\cite{seglosssurvey} is applied as mask loss:
\begin{equation}
\mathcal{L}_{\text{m}} = -(m^{\text{ref}} \log m + (1 - m^{\text{ref}}) \log (1 - m)),
\end{equation}
where $m$ denotes the object mask. A shift and scale invariant depth loss is utilized to guide geometry:
\begin{equation}
\mathcal{L}_{\text{d}} = \|D^* - D_{\text{pred}}^*\|_1,
\end{equation}
where $D^*$ and $D_{\text{pred}}^*$ are per-frame rendered depths and monocularly estimated depths~\cite{zoedepth} respectively.
The depth values are computed following a normalization strategy~\cite{ranftl2020towards}:
\begin{equation}
D^* = \frac{D - \text{median}(D)}{\frac{1}{M} \sum_{i=1}^{M} |D - \text{median}(D)|},
\end{equation}
where $M$ denotes the number of valid pixels. 
The overall loss combines these components:
\begin{equation}
\mathcal{L}_{\text{ref}}=(1-\lambda_{\mathrm{SSIM}})\mathcal{L}_{1}+\lambda_{\mathrm{SSIM}}\mathcal{L}_{\mathrm{D-SSIM}}+\lambda_{\mathrm{m}}\mathcal{L}_{\mathrm{m}}+\lambda_{\mathrm{d}}\mathcal{L}_{\mathrm{d}},
\end{equation}
where $\lambda_{\mathrm{SSIM}}$, $\lambda_{\mathrm{m}}$, and $\lambda_{\mathrm{d}}$ control the magnitude of each term. Thanks to the efficient initialization, our training speed is remarkably fast. It only takes 1 minute to train a coarse Gaussian representation $\mathcal{G}_c$ at a resolution of 779 $\times$ 520.

\begin{figure}[t]
  \centering
  \includegraphics[width=1.0\linewidth]{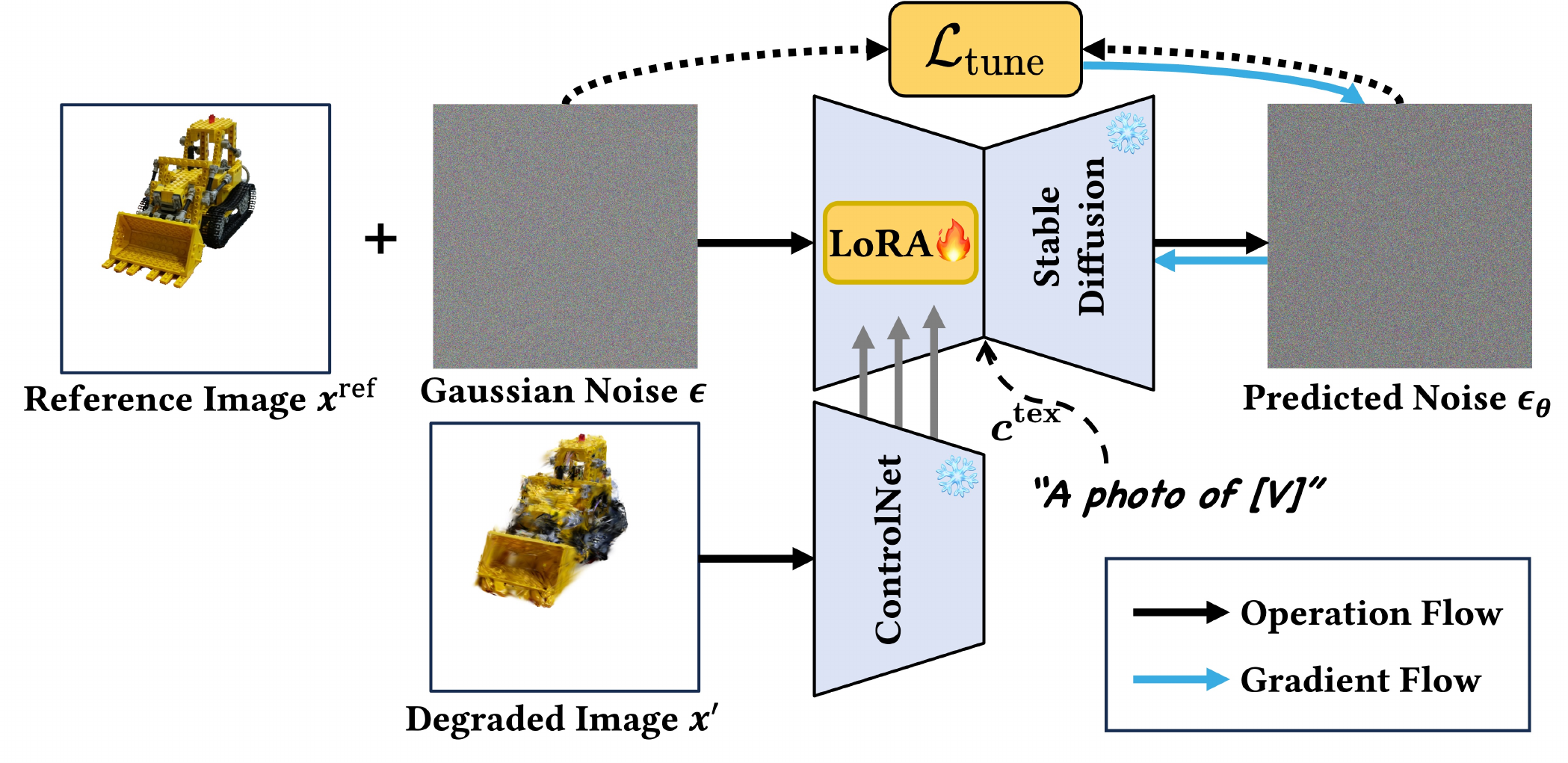}
  \vspace{-2.5em}
  \caption{Illustration of Gaussian repair model setup. First, we add Gaussian noise $\epsilon$ to a reference image $x^{\text{ref}}$ to form a noisy image. Next, this noisy image along with $x^{\text{ref}}$'s corresponding degraded image $x^\prime$ are passed to a pre-trained fixed ControlNet with learnable LoRA layers to predict a noise distribution $\epsilon_{\theta}$. We use the differences among $\epsilon$ and $\epsilon_{\theta}$ to fine-tune the parameters in LoRA layers.}
  \Description{}
  \label{fig: repair_model illustration}
  \vspace{-1.5em}
\end{figure}

\subsection{Gaussian Repair Model Setup} \label{Method: Gaussian Repair Model Setup}
Combining visual hull initialization and floater elimination significantly enhances 3DGS performance for NVS in sparse 360\deg contexts. While the fidelity of our reconstruction is generally passable, $\mathcal{G}_c$ still suffers in regions that are poorly observed, regions with occlusion, or even unobserved regions. These challenges loom over the completeness of the reconstruction, like the sword of Damocles.

To mitigate these issues, we introduce a Gaussian repair model $\mathcal{R}$ designed to correct the aberrant distribution of $\mathcal{G}_c$. Our $\mathcal{R}$ takes corrupted rendered images $x^\prime(\mathcal{G}_c, \pi^{\text{nov}})$ as input and outputs photo-realistic and high-fidelity images $\hat{x}$. This image repair capability can be used to refine the 3D Gaussians, leading to learning better structure and appearance details.

Sufficient data pairs are essential for training $\mathcal{R}$ but are rare in existing datasets. To this end, we adopt two main strategies for generating adequate image pairs, \ie, \textbf{leave-one-out training} and \textbf{adding 3D noises}. 
For leave-one-out training, we build $N$ subsets from the $N$ input images, each containing $N-1$ reference images and $1$ left-out image $x^{\text{out}}$.
Then we train $N$ 3DGS models with reference images of these subsets, termed as $\{\mathcal{G}_c^i\}_{i=0}^{N-1}$. 
After specific iterations, we use the left-out image $x^{\text{out}}$ to continue training each Gaussian model $\{\mathcal{G}_c^i\}_{i=0}^{N-1}$ into $\{\hat{\mathcal{G}}_c^i\}_{i=0}^{N-1}$. Throughout this process, the rendered images from the left-out view at different iterations are stored to form the image pairs along with left-out image $x^{\text{out}}$ for training the repair model. 
Note that training these left-out models costs little, with less than $N$ minutes in total.
The other strategy is to add 3D noises $\epsilon_s$ onto Gaussian attributes. The $\epsilon_s$ are derived from the mean $\mu_{\Delta}$ and variance $\sigma_{\Delta}$ of attribute differences between $\{\mathcal{G}_c^i\}_{i=0}^{N-1}$ and $\{\hat{\mathcal{G}}_c^i\}_{i=0}^{N-1}$. 
This allows us to render more degraded images $x^\prime(\mathcal{G}_c(\epsilon_s), \pi^{\text{ref}})$ at all reference views from the created noisy Gaussians, resulting in extensive image pairs $(X^\prime, X^{\text{ref}})$.

We inject LoRA weights and fine-tune a pre-trained ControlNet~\cite{controlnet_v11} using the generated image pairs as our Gaussian repair model. The training procedure is shown in Fig.~\ref{fig: repair_model illustration}. The loss function, based on Eq.~\ref{eq: controlnet loss}, is defined as:
\begin{equation}
    \mathcal{L}_{\text{tune}} = \mathbb{E}_{x^{\text{ref}}, t, \epsilon, x^\prime} \left[ \| (\epsilon_{\theta}(x^{\text{ref}}_t, t, x^\prime, c^{\text{tex}}) - \epsilon) \|_2^2 \right],
\end{equation}
where $c^{\text{tex}}$ denotes an object-specific language prompt, defined as ``a photo of [V],'' as per Dreambooth~\cite{dreambooth}. Specifically, we inject LoRA layers into the text encoder, image condition branch, and U-Net for fine-tuning. Please refer to the Appendix for details.

\subsection{Gaussian Repair with Distance-Aware Sampling} \label{Method: Gaussian Repair with Distance-Aware Sampling}
After training $\mathcal{R}$, we distill its target object priors into $\mathcal{G}_c$ to refine its rendering quality. 
The object information near the reference views is abundant.
This observation motivates designing distance as a criterion in identifying views that need rectification, leading to distance-aware sampling.  

\begin{figure}[t]
  \centering
  \includegraphics[width=0.65\linewidth]{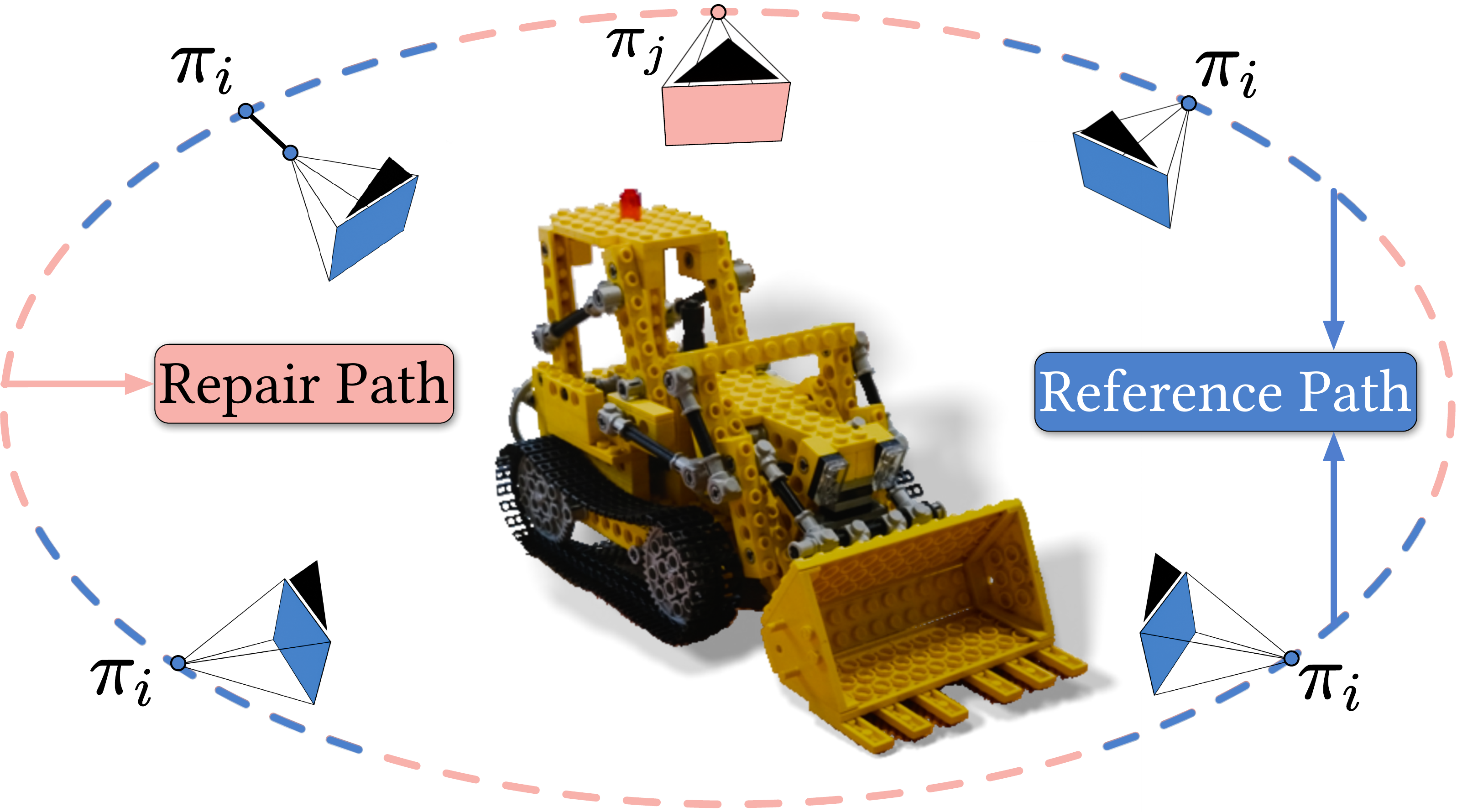}
  \vspace{-1em}
  \caption{Illustration of our distance-aware sampling. Blue and red indicate the reference and repair path, respectively.}
  \Description{}
  \label{fig: distance}
  \vspace{-1em}
\end{figure}

Specifically, we establish an elliptical path aligned with the training views and focus on a central point. Arcs near $\Pi^{ref}$, where we assume $\mathcal{G}_c$ renders high-quality images, form the reference path. The other arcs, yielding renderings, need to be rectified and define the repair path, as depicted in Fig.~\ref{fig: distance}. In each iteration, novel viewpoints, $\pi_j \in \Pi^{\text{nov}}$, are randomly sampled among the repair path. For each $\pi_j$, we render the corresponding image $x_j(\mathcal{G}_c, \pi_j)$, encode it to be $\mathcal{E}(x_j)$ by the latent diffusion encoder $\mathcal{E}$ and pass $\mathcal{E}(x_j)$ to the image conditioning branch of $\mathcal{R}$. Simultaneously, a cloned $\mathcal{E}(x_j)$ is disturbed into a noisy latent \(z_t\):
\begin{align}
z_t = \sqrt{\bar{\alpha}_t}\mathcal{E}(x_j)+\sqrt{1-\bar{\alpha}_t}\epsilon,
\text{ where }\epsilon\sim\mathcal{N}(0,I), t\in [0, T],
\end{align}
which is similar to SDEdit~\cite{sdedit}. 
We then generate a sample \(\hat{x}_j\) from $\mathcal{R}$ by running DDIM sampling~ \cite{ddim} over \(k = \lfloor 50 \cdot \frac{t}{T} \rfloor \) steps and forwarding the diffusion decoder $\mathcal{D}$:
\begin{equation}
    \hat{x}_j = \mathcal{D}(\mathrm{DDIM}(z_t, \mathcal{E}(x_j))),
\end{equation}
where $\mathcal{E}$ and $\mathcal{D}$ are from the VAE model used by the diffusion model. The distances from $\pi_j$ to $\Pi^{ref}$ is used to weight the reliability of \(\hat{x}_j\), guiding the optimization with a loss function:
\begin{align}
\mathcal{L}_{\text{rep}} = \mathbb{E}_{\pi_j,t} \bigl[ w(t) \lambda(\pi_j) \bigl(&\|x_j - \hat{x}_j\|_1 + \|x_j - \hat{x}_j\|_2 \notag + L_p(x_j, \hat{x}_j) \bigr) \bigr], \notag \\
\text{where } \lambda(\pi_j) &= \frac{2 \cdot \min_{i=1}^{N}(\|\pi_j - \pi_i\|_2)}{d_{\text{max}}}.
\end{align}
Here, $L_p$ denotes the perceptual similarity metric LPIPS~\cite{lpips}, $w(t)$ is a noise-level modulated weighting function from DreamFusion~\cite{dreamfusion}, $\lambda(\pi_j)$ denotes a distance-based weighting function, and $d_{\text{max}}$ is the maximal distance among neighboring reference viewpoints. 
To ensure coherence between 3D Gaussians and reference images, we continue training $\mathcal{G}_c$ with $\mathcal{L}_{\text{ref}}$ during the whole Gaussian repair procedure.

\subsection{COLMAP-Free GaussianObject (CF-GaussianObject)} \label{Method: COLMAP-free GaussianObject}
Current SOTA sparse view reconstruction methods rely on precise camera parameters, including intrinsics and poses, obtained through an SfM pipeline with dense input, limiting their usability in daily applications. This process can be cumbersome and unreliable in sparse-view scenarios where matched features are insufficient for accurate reconstruction.

To overcome this limitation, we introduce an advanced sparse matching model, DUSt3R \cite{dust3r}, into GaussianObject to enable COLMAP-free sparse 360\deg reconstruction. Given reference input images \(X^{\text{ref}}\), DUSt3R is formulated as:
\begin{equation}
    \mathcal{P}, \hat{\Pi}^{\text{ref}}, \hat{K}^{\text{ref}} = \text{DUSt3R}(X^{\text{ref}}),
\end{equation}
where \(\mathcal{P}\) is an estimated coarse point cloud of the scene, and $\hat{\Pi}^{\text{ref}}$, $\hat{K}^{\text{ref}}$ are the predicted camera poses and intrinsics of \(X^{\text{ref}}\), respectively. For CF-GaussianObject, we modify the intrinsic recovery module within DUSt3R, allowing \(x_i \in X^{\text{ref}}\) to share the same intrinsic $\hat K$. This adaption enables the retrieval of $\mathcal{P}$, \(\hat{\Pi}^{\text{ref}}\), and $\hat K$. Besides, we apply structural priors with a visual hull to \(\mathcal{P}\) to initialize 3D Gaussians. After initialization, we optimize $\hat{\Pi}^{\text{ref}}$ and the initialized 3D Gaussians using \(X^{\text{ref}}\) and depth maps rendered from $\mathcal{P}$ simultaneously. Besides, we introduce a regularization loss to constrain deviations from $\hat{\Pi}^{\text{ref}}$, enhancing the robustness of the optimization. 
After optimization, the 3D Gaussians and camera parameters are used for constructing the Gaussian repair model and Gaussian repairing process as described in Sec.~\ref{Method: Gaussian Repair Model Setup} and Sec.~\ref{Method: Gaussian Repair with Distance-Aware Sampling}. Refer to the Appendix for more details.

\begin{figure*}[htbp]
\centering
\includegraphics[width=0.95\textwidth]{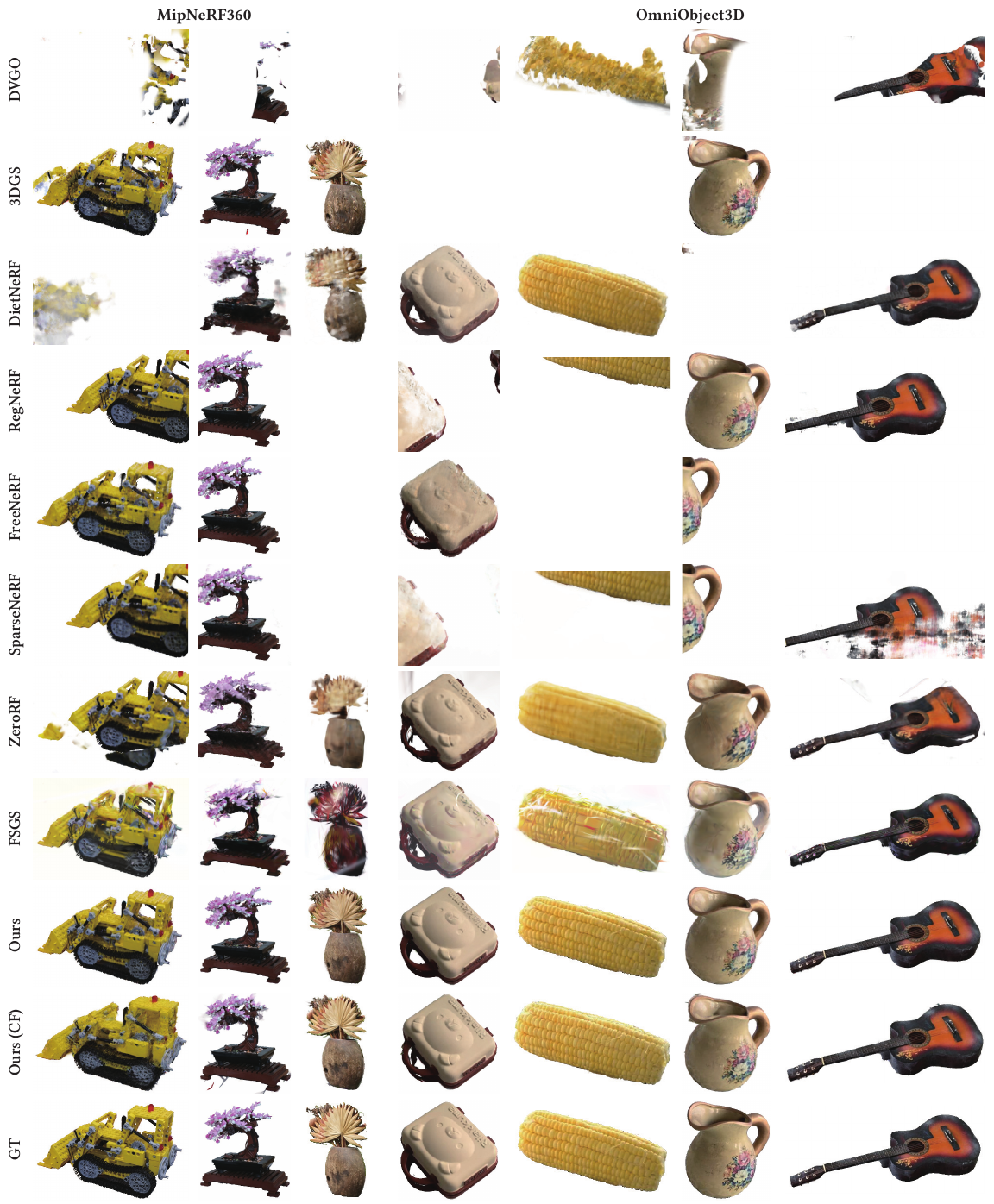}
\caption{Qualitative examples on the MipNeRF360 and OmniObject3D dataset with 4 input views. Many methods fail to reach a coherent 3D representation, resulting in floaters and disjoint pixel patches. 
A pure white image indicates a total miss of the object by the corresponding method, usually caused by overfitting the input images.
}
\Description{}
\label{fig: qualitative}
\end{figure*}

\begin{table*}[thbp]
\newcommand{\xmark}{\ding{55}}
\centering
\small
\caption{Comparisons with varying input views. $\text{LPIPS}^* = \text{LPIPS}\times 10^2$ throughout this paper. Best results are highlighted as \colorbox{metric-1}{1st}, \colorbox{metric-2}{2nd} and \colorbox{metric-3}{3rd}.}
\vspace{-1em}
\label{tab: compare}
\begin{tabular}{c|l|ccc|ccc|ccc}
\toprule
\multirow{2}{*}{} & \multirow{2}{*}{Method} & \multicolumn{3}{c|}{4-view} & \multicolumn{3}{c|}{6-view} & \multicolumn{3}{c}{9-view} \\
 & & \textbf{LPIPS$^*$} $\downarrow$ & \textbf{PSNR} $\uparrow$ & \textbf{SSIM} $\uparrow$ & \textbf{LPIPS$^*$} $\downarrow$ & \textbf{PSNR} $\uparrow$ & \textbf{SSIM} $\uparrow$ & \textbf{LPIPS$^*$} $\downarrow$ & \textbf{PSNR} $\uparrow$ & \textbf{SSIM} $\uparrow$ \\ 
\midrule
\multirow{6}{*}[-2.7em]{\rotatebox[origin=c]{90}{MipNeRF360}}
 & DVGO~\cite{dvgo}             & 24.43                     & 14.39                     & 0.7912                     & 26.67                     & 14.30                     & 0.7676                     & 25.66                     & 14.74                     & 0.7842                     \\
 & 3DGS~\cite{3dgs}             & 10.80                     & 20.31                  
   & 0.8991                     & 8.38                     & 22.12                     & 0.9134                     & 6.42                     & 24.29                     & 0.9331                     \\
 & DietNeRF~\cite{dietnerf}     & 11.17                     & 18.90                     & 0.8971                     & 6.96                     & 22.03                     & \cellcolor{metric-3}0.9286 & 5.85                     & 23.55                     & 0.9424                     \\
 & RegNeRF~\cite{regnerf}       & 20.44                     & 13.59                     & 0.8476                     & 20.72                     & 13.41                     & 0.8418                     & 19.70                     & 13.68                     & 0.8517                     \\
 & FreeNeRF~\cite{freenerf}     & 16.83                     & 13.71                     & 0.8534                     & \cellcolor{metric-3}6.84 & 22.26                     & \cellcolor{metric-2}0.9332 & 5.51                     & \cellcolor{metric-3}27.66 & \cellcolor{metric-2}0.9485 \\
 & SparseNeRF~\cite{sparsenerf} & 17.76                     & 12.83                     & 0.8454                     & 19.74                     & 13.42                     & 0.8316                     & 21.56                     & 14.36                     & 0.8235                     \\
 & ZeroRF~\cite{zerorf}         & 19.88                     & 14.17                     & 0.8188                     & 8.31                     & \cellcolor{metric-2}24.14 & 0.9211                     & \cellcolor{metric-2}5.34 & \cellcolor{metric-2}27.78 & \cellcolor{metric-3}0.9460 \\
 & FSGS~\cite{fsgs}             & \cellcolor{metric-3}9.51 & \cellcolor{metric-3}21.07 & \cellcolor{metric-2}0.9097 & 7.69                     & 22.68             
        & 0.9264                     & 6.06                     & 25.31                     & 0.9397                     \\
\cmidrule{2-11}
 & \name (Ours)                 & \cellcolor{metric-1}4.98 & \cellcolor{metric-1}24.81 & \cellcolor{metric-1}0.9350 & \cellcolor{metric-1}3.63 & \cellcolor{metric-1}27.00     & \cellcolor{metric-1}0.9512 & \cellcolor{metric-1}2.75 & \cellcolor{metric-1}28.62 & \cellcolor{metric-1}0.9638 \\
 & CF-\name (Ours)                    & \cellcolor{metric-2}8.47 & \cellcolor{metric-2}21.39 & \cellcolor{metric-3}0.9014 & \cellcolor{metric-2}5.71 & \cellcolor{metric-3}24.06 & 
 0.9269                     & \cellcolor{metric-3}5.50 & 24.39                     & 0.9300 \\
\midrule
\multirow{6}{*}[-2.5em]{\rotatebox[origin=c]{90}{OmniObject3D}}
 & DVGO~\cite{dvgo}             & 14.48                     & 17.14                     & 0.8952                     & 12.89                     & 18.32                     & 0.9142                     & 11.49                     & 19.26                     & 0.9302                     \\
 & 3DGS~\cite{3dgs}             & 8.60                     & 17.29                     & 0.9299                     & 7.74                     & 18.29                     & 0.9378                     & 6.50                     & 20.26                     & 0.9483                     \\
 & DietNeRF~\cite{dietnerf}     & 11.64                     & 18.56                     & 0.9205                     & 10.39                     & 19.07                     & 0.9267                     & 10.32                     & 19.26                     & 0.9258                     \\
 & RegNeRF~\cite{regnerf}       & 16.75                     & 15.20                     & 0.9091                     & 14.38                     & 15.80                     & 0.9207                     & 10.17                     & 17.93                     & 0.9420                     \\
 & FreeNeRF~\cite{freenerf}     & 8.28                     & 17.78                     & 0.9402                     & 7.32                     & 19.02                     & 0.9464                     & 7.25                     & 20.35                     & 0.9467                     \\
 & SparseNeRF~\cite{sparsenerf} & 17.47                     & 15.22                     & 0.8921                     & 21.71                     & 15.86                     & 0.8935                     & 23.76                     & 17.16                     & 0.8947                     \\
 & ZeroRF~\cite{zerorf}         & \cellcolor{metric-3}4.44 & \cellcolor{metric-3}27.78 & \cellcolor{metric-3}0.9615 & \cellcolor{metric-3}3.11 & \cellcolor{metric-2}31.94 & \cellcolor{metric-3}0.9731 & \cellcolor{metric-3}3.10 & \cellcolor{metric-2}32.93 & \cellcolor{metric-3}0.9747 \\
 & FSGS~\cite{fsgs}             & 6.25                     & 24.71                   
  & 0.9545                     & 6.05                     & 26.36                     & 0.9582                     & 4.17                     & 29.16                   
  & 0.9695                     \\
\cmidrule{2-11}
 & \name (Ours)                 & \cellcolor{metric-1}2.07 & \cellcolor{metric-1}30.89 & \cellcolor{metric-1}0.9756 & \cellcolor{metric-1}1.55 & \cellcolor{metric-1}33.31 & \cellcolor{metric-1}0.9821 & \cellcolor{metric-1}1.20 & \cellcolor{metric-1}35.49 & \cellcolor{metric-1}0.9870 \\
 & CF-\name (Ours)                   & \cellcolor{metric-2}2.62 & \cellcolor{metric-2}28.51 & \cellcolor{metric-2}0.9669 & \cellcolor{metric-2}2.03 & \cellcolor{metric-3}30.73 & \cellcolor{metric-2}0.9738 & \cellcolor{metric-2}2.08 & \cellcolor{metric-3}31.23 & \cellcolor{metric-2}0.9757 \\
\bottomrule
\end{tabular}
\end{table*}

\begin{table}[t!]
\setlength{\tabcolsep}{3.2pt}
\centering
\caption{Quantitative comparisons on the OpenIllumination dataset. Methods with \textdagger\ means the metrics are from the ZeroRF paper~\cite{zerorf}.}
\vspace{-1em}
\small
\label{tab: compare 2}
\begin{tabular}{l|ccc|ccc}
\toprule
\multirow{2}{*}{Method} & \multicolumn{3}{c|}{4-view} & \multicolumn{3}{c}{6-view} \\
 & \textbf{LPIPS$^*$} $\downarrow$ & \textbf{PSNR} $\uparrow$ & \textbf{SSIM} $\uparrow$ & \textbf{LPIPS$^*$} $\downarrow$ & \textbf{PSNR} $\uparrow$ & \textbf{SSIM} $\uparrow$ \\ 
\midrule
DVGO                   & 11.84                     & 21.15                     & 0.8973                     & \cellcolor{metric-3}8.83 & 23.79                     & 0.9209                     \\
3DGS                   & 30.08                     & 11.50                     & 0.8454                     & 29.65                     & 11.98                     & 0.8277                     \\
DietNeRF$^\dagger$ & \cellcolor{metric-3}10.66 & \cellcolor{metric-3}23.09 & \cellcolor{metric-1}0.9361 & 9.51                     & \cellcolor{metric-3}24.20 & \cellcolor{metric-3}0.9401 \\
RegNeRF$^\dagger$   & 47.31                     & 11.61                     & 0.6940                     & 30.28                     & 14.08                     & 0.8586                     \\
FreeNeRF$^\dagger$ & 35.81                     & 12.21                     & 0.7969                     & 35.15                     & 11.47                     & 0.8128                     \\
SparseNeRF      & 22.28                     & 13.60                     & 0.8808                     & 26.30                     & 12.80                     & 0.8403                     \\
ZeroRF$^\dagger$     & \cellcolor{metric-2}9.74 & \cellcolor{metric-2}24.54 & \cellcolor{metric-3}0.9308 & \cellcolor{metric-2}7.96 & \cellcolor{metric-2}26.51 & \cellcolor{metric-2}0.9415 \\
Ours                              & \cellcolor{metric-1}6.71 & \cellcolor{metric-1}24.64 & \cellcolor{metric-2}0.9354 & \cellcolor{metric-1}5.44 & \cellcolor{metric-1}26.54 & \cellcolor{metric-1}0.9443 \\
\bottomrule
\end{tabular}
\vspace{-1em}
\end{table}

\begin{figure}[t]
\centering
\includegraphics[width=\linewidth]{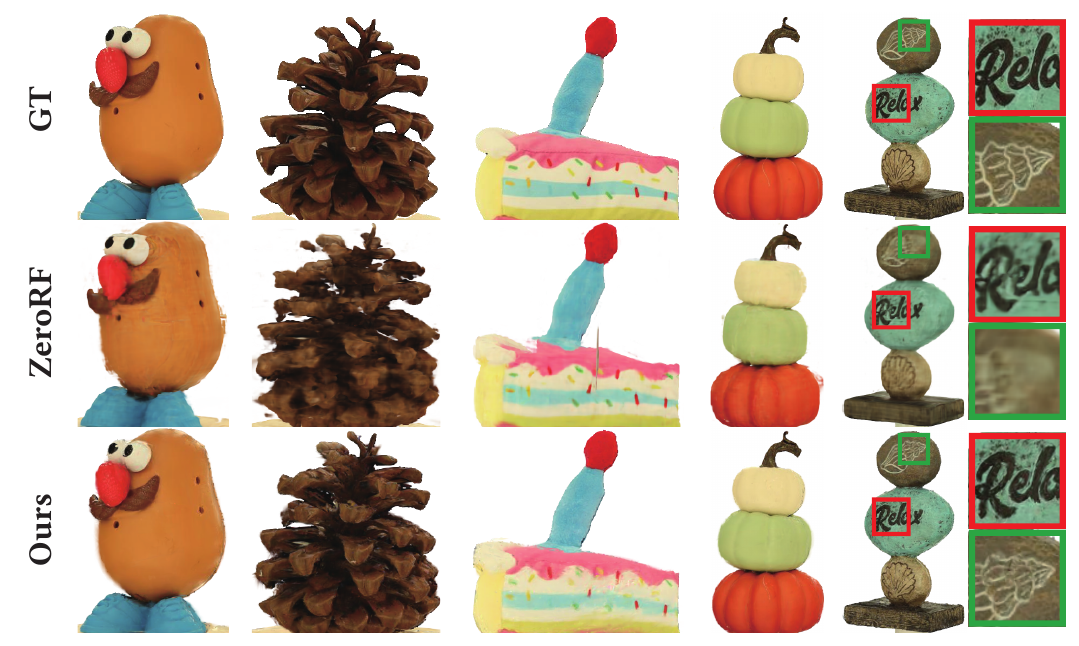}
\vspace{-1em}
\caption{Qualitative results on the OpenIllumination dataset. Although ZeroRF shows competitive PSNR and SSIM, its renderings often appear blurred. While \name outperforms in restoring fine details, achieving a significant perceptual quality advantage.}
\Description{}
\label{fig: qualitative results2}
\end{figure}

\section{Experiments}
\subsection{Implementation Details}
\label{subsec: implem}

Our framework, illustrated in Fig.~\ref{fig: framework}, is based on 3DGS~\cite{3dgs} and threestudio~\cite{threestudio2023}. 
The 3DGS model is trained for 10k iterations in the initial optimization, with periodic floater elimination every 500 iterations. The monocular depth for $\mathcal{L}_{\text{d}}$ is predicted by ZoeDepth~\cite{zoedepth}. We use a ControlNet-Tile~\cite{controlnet_v11} model based on stable diffusion v1.5~\cite{ldm} as our repair model's backbone. LoRA~\cite{lora} weights, injected into the text-encoder and transformer blocks using minLoRA~\cite{minlora}, are trained for 1800 steps at a LoRA rank of 64 and a learning rate of $10^{-3}$. $\mathcal{G}_c$ is trained for another 4k iterations during distance-aware sampling. For the first 2800 iterations, optimization involves both a reference image and a repaired novel view image, with the weight of $\mathcal{L}_{\text{rep}}$ progressively decayed from $1.0$ to $0.1$. The final 1200-step training only involves reference views. The whole process of \name takes about 30 minutes on a GeForce RTX 3090 GPU for 4 input images at a 779 × 520 resolution. For more details, please refer to the Appendix.

\subsection{Datasets} \label{subsec: Datasets}
We evaluate \name on three datasets suited for sparse-view 360\deg object reconstruction with varying input views, including Mip-NeRF360~\cite{mip360}, OmniObject3D~\cite{OmniObject3D}, and OpenIllumination~\cite{OpenIllumination}. Additionally, we use an iPhone 13 to capture four views of some daily-life objects to show the COLMAP-free performance. SAM~\cite{sam} is used to obtain masks of the target objects.

\begin{figure}[t]
\centering
\includegraphics[width=1.0\linewidth]{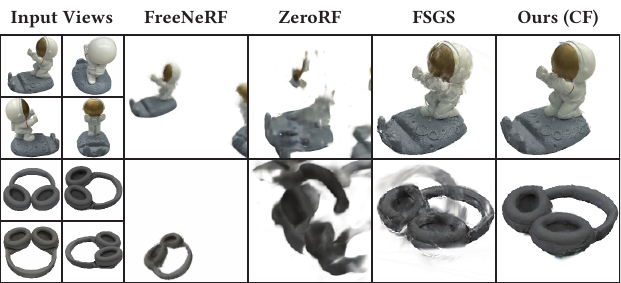}
\vspace{-2em}
\caption{Qualitative results on our-collected images captured by an iPhone 13. We equip other SOTAs with camera parameters predicted by DUSt3R for fair comparison. The results demonstrate the superior performance of our CF-\name among casually captured images, with fine details and higher visual quality.}
\Description{}
\label{fig: cf_compare}
\end{figure}

\subsection{Evaluation} \label{subsec: eval}
\rev{\textit{Sparse 360\deg Reconstruction Performance.}}
We evaluate the performance of \name against several reconstruction baselines, including the vanilla 3DGS~\cite{3dgs} with random initialization and DVGO~\cite{dvgo}, and various few-view reconstruction models on the three datasets. 
Compared methods of RegNeRF~\cite{regnerf}, DietNeRF~\cite{dietnerf}, SparseNeRF~\cite{sparsenerf}, and ZeroRF~\cite{zerorf} utilize a variety of regularization techniques. Besides, FSGS~\cite{fsgs} is also built upon Gaussian splatting with SfM-point initialization. Note that we supply extra SfM points to FSGS so that it can work with the highly sparse 360\deg setting. 
Since camera pose estimation often suffers from scale and positional errors compared to ground truth, we adopt the evaluation used for COLMAP-free methods under dense view settings~\cite{fu2023colmap,wang2021nerf}.
All models are trained using publicly released codes.

Table~\ref{tab: compare} and~\ref{tab: compare 2} present the view-synthesis performance of \name compared to existing methods on the MipNeRF360, OmniObject3D, and OpenIllumination datasets. Experiments show that \name consistently achieves SOTA results in all datasets, especially in the perceptual quality -- LPIPS\allowbreak. Although \name is designed to address extremely sparse input views, it still outperforms other methods with more input views, \textit{i.e.} 6 and 9, further proving the effectiveness.
Notably, \name excels with as few as 4 views and significantly improves LPIPS over FSGS from 0.0951 to 0.0498 on MipNeRF360. This improvement is critical, as LPIPS is a key indicator of perceptual quality~\cite{hypernerf}.

Fig.~\ref{fig: qualitative} and Fig.~\ref{fig: qualitative results2} illustrate rendering results of various methods across different datasets with only 4 input views. We observe that \name achieves significantly better visual quality and fidelity than the competing models. 
We find that implicit representation based methods and random initialized 3DGS fail in extremely sparse settings, typically reconstructing objects as fragmented pixel patches. This confirms the effectiveness of integrating structure priors with explicit representations. 
Although ZeroRF exhibits competitive PSNR and SSIM on OpenIllumination, its renderings are blurred and lack details, as shown in Fig.~\ref{fig: qualitative results2}. In contrast, \name demonstrates fine-detailed reconstruction. This superior perceptual quality highlights the effectiveness of the Gaussian repair model. 
It is highly suggested to refer to comprehensive video comparisons included in supplementary materials. 

\rev{\textit{Comparison with LRMs.}}
\rev{We further compare \name to recently popular LRM-like feed-forward reconstruction methods, \ie LGM~\cite{lgm} and TriplaneGaussian (TGS)~\cite{tgs} which are publicly available. The comparisons are shown in Table~\ref{tab: lrm} on the challenging MipNeRF360 dataset. Given that TriplaneGaussian accommodates only a single image input, we feed it with frontal views of objects. 
LGM requires placing the target object at the world coordinate origin with cameras oriented towards it at an elevation of 0\deg and azimuths of 0\rev{$^\circ$}, 90\rev{$^\circ$}, 180\rev{$^\circ$}, and 270\rev{$^\circ$}. Therefore, we report two versions of LGM -- LGM-4 which uses four sparse captures as input views directly, and LGM-1 which uses MVDream~\cite{mvdream} to generate images that comply with LGM's setup requirements following its original manner.
Results show that the strict requirements among input views significantly hinder the sparse reconstruction performance of LRM-like models with in-the-wild captures. In contrast, \name does not require extensive pre-training, has no restrictions on input views, and can reconstruct any complex object in daily life.}

\begin{figure}[t]
\centering
\includegraphics[width=\linewidth]{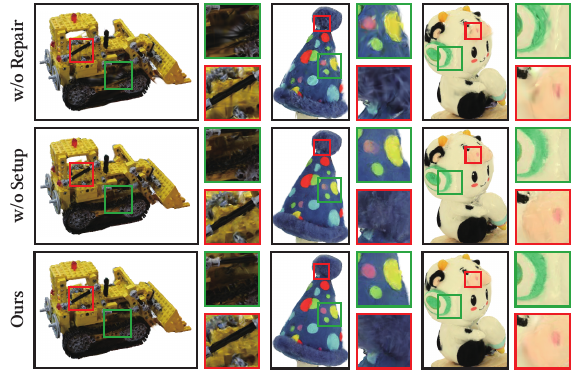}
\vspace{-1em}
\caption{Importance of our Gaussian repair model setup. Without the Gaussian repair process or the finetuning of the ControlNet, the renderings exhibit noticeable artifacts and lack of details, particularly in areas with insufficient view coverage. Zoom in for better comparison.}
\Description{}
\label{fig: qualitative result gaussian repair}
\end{figure}

\rev{\textit{Performance of CF-GaussianObject.}}
CF-GaussianObject is evaluated on the MipNeRF360 and OmniObject3D datasets, with results detailed in Table~\ref{tab: compare} and Fig.~\ref{fig: qualitative}. Though CF-GaussianObject exhibits some performance degradation, it eliminates the need for precise camera parameters, significantly enhancing its practical utility. Its performance remains competitive compared to other SOTA methods that depend on accurate camera parameters. Notably, we observe that the performance degradation correlates with an increase in the number of input views, primarily due to declines in the accuracy of DUSt3R's estimates as the number of views rises. As demonstrated in Fig.~\ref{fig: cf_compare}, comparative experiments on smartphone-captured images confirm the superior reconstruction capabilities and visual quality of CF-GaussianObject. More visualization of CF-GaussianObject can be found in our appendix and supplementary materials.

\subsection{Ablation Studies} \label{subsec: abla}
\begin{figure}[t]
\centering
\includegraphics[width=\linewidth]{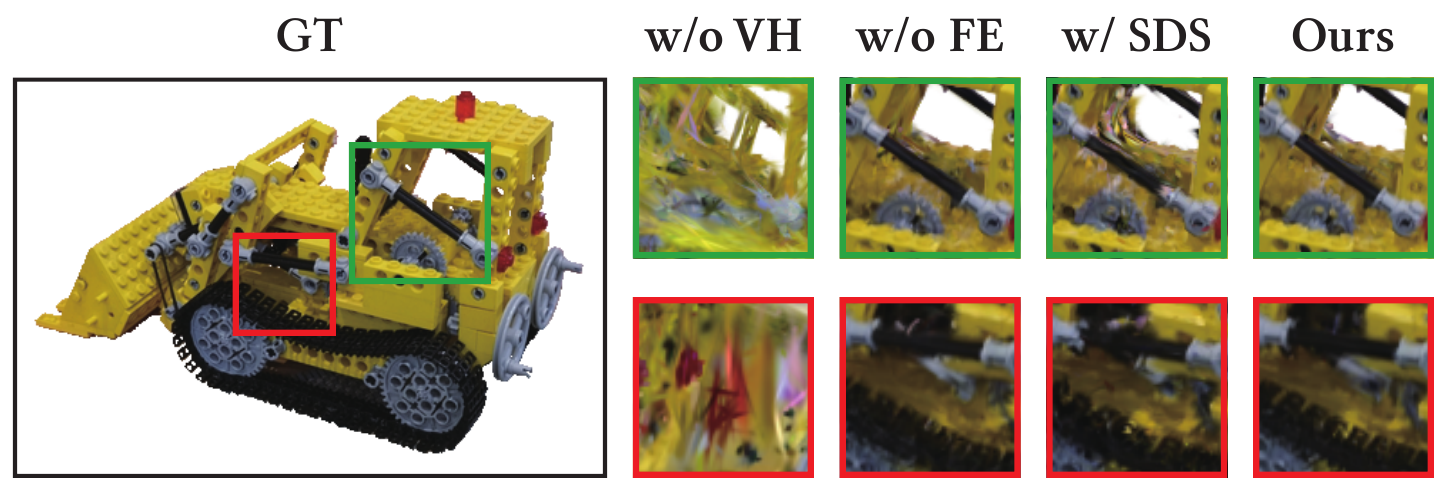}
\vspace{-2em}
\caption{Ablation study on different components. ``VH'' denotes for visual hull and ``FE'' is floater elimination. The ``GT'' image is from a test view.}
\Description{}
\label{fig: ablation components}
\end{figure}

\textit{Key Components.}
We conduct a series of experiments to validate the effectiveness of each component. The following experiments are performed on MipNeRF360 with 4 input views, and averaged metric values are reported. We disable visual hull initialization, floater elimination, Gaussian repair model setup, and Gaussian repair process once at a time to verify their effectiveness. The Gaussian repair loss is further compared with the Score Distillation Sampling (SDS) loss~\cite{dreamfusion}, and the depth loss is ablated. 
The results, presented in Table~\ref{tab: ablation components} and Fig.~\ref{fig: ablation components}, indicate that each element significantly contributes to performance, with their absence leading to a decline in results. Particularly, omitting visual hull initialization results in a marked decrease in performance.
Gaussian repair model setup and the Gaussian repair process significantly enhance visual quality, and the absence of either results in a substantial decline in perceptual quality as shown in Fig.~\ref{fig: qualitative result gaussian repair}. 
Contrary to its effectiveness in text-to-3D or single image-to-3D, SDS results in unstable optimization and diminished performance in our context. The depth loss shows marginal promotion, mainly for LPIPS and SSIM. We apply it to enhance the robustness of our framework.

\begin{table}[t]
\centering
\caption{\rev{Quantitative comparisons with LRM-like methods on MipNeRF360.}}
\vspace{-1em}
\setlength{\tabcolsep}{7pt}
\label{tab: lrm}
\begin{tabular}{l|ccc}
\toprule
Method & \textbf{LPIPS$^*$} $\downarrow$ & \textbf{PSNR} $\uparrow$ & \textbf{SSIM} $\uparrow$ \\
\midrule
TGS~\cite{tgs}  & 9.14 & 18.07 & 0.9073 \\
LGM-4~\cite{lgm}           & 9.20 & 17.97 & 0.9071 \\
LGM-1~\cite{lgm}               & 9.13 & 17.46 & 0.9071 \\
\midrule
\name (Ours)          & \textbf{4.99} & \textbf{24.81} & \textbf{0.9350} \\
\bottomrule
\end{tabular}
\vspace{-1em}
\end{table}

\begin{figure}[t]
\centering
\includegraphics[width=1.0\linewidth]{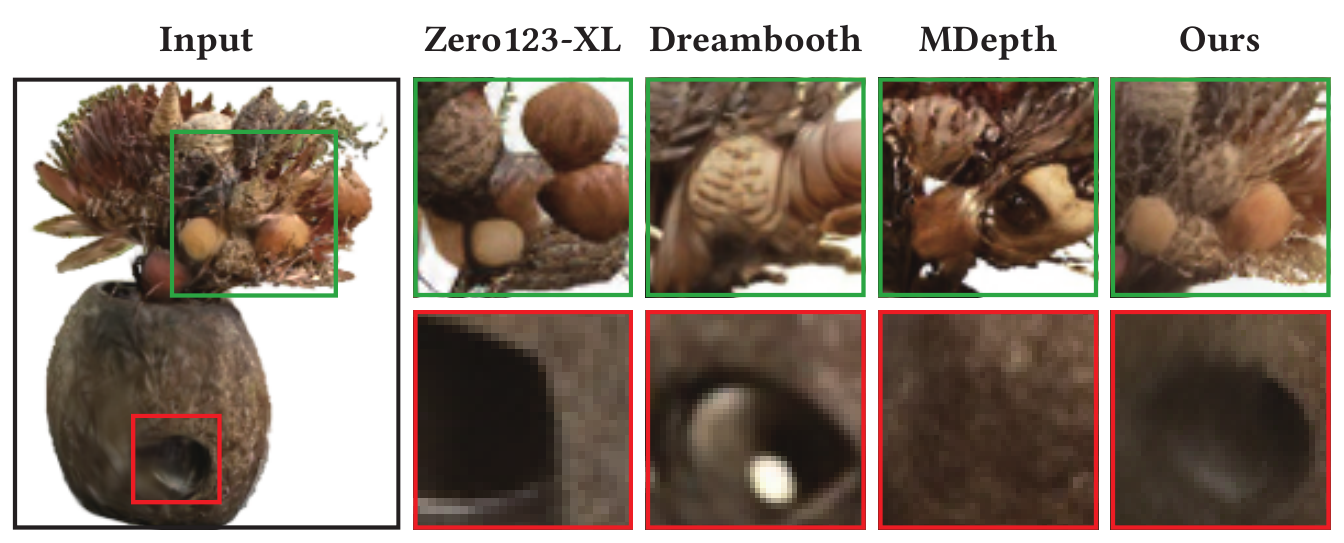}
\vspace{-2em}
\caption{Qualitative comparisons by ablating different Gaussian repair model setup methods. ``MDepth'' denotes the repair model with masked monocular depth estimation as the condition.}
\Description{}
\label{fig: repair model structure}
\end{figure}

\begin{table}[t]
\newcommand{\xmark}{\ding{55}}
\centering
\caption{Ablation study on key components.}
\vspace{-1em}
\setlength{\tabcolsep}{4pt}
\label{tab: ablation components}
\begin{tabular}{l|ccc}
\toprule
Method & \textbf{LPIPS$^*$} $\downarrow$ & \textbf{PSNR} $\uparrow$ & \textbf{SSIM} $\uparrow$ \\
\midrule
Ours w/o Visual Hull          & 12.72          & 15.95          & 0.8719          \\
Ours w/o Floater Elimination  & 4.99           & 24.73          & 0.9346          \\
Ours w/o Setup           & 5.53           & 24.28          & 0.9307          \\
Ours w/o Gaussian Repair      & 5.55          & 24.37          & 0.9297          \\
Ours w/o Depth Loss           & 5.09          & \textbf{24.84} & 0.9341          \\
\midrule
Ours w/ SDS \cite{dreamfusion} & 6.07          & 22.42          & 0.9188          \\
\midrule
\name (Ours)                   & \textbf{4.98} & 24.81          & \textbf{0.9350} \\
\bottomrule
\end{tabular}
\vspace{-1em}
\end{table}

\textit{Structure of Repair Model.}
Our repair model is designed to generate photo-realistic and 3D-consistent views of the target object. This is achieved by leave-one-out training and perturbing the attributes of \gs to create image pairs for fine-tuning a pre-trained image-conditioned ControlNet. Similarities can be found in Dreambooth~\cite{dreambooth}, which aims to generate specific subject images from limited inputs.
To validate the efficacy of our design, we evaluate the samples generated by our Gaussian repair model and other alternative structures. 
The first is implemented with Dreambooth~\cite{dreambooth,dreambooth3d}, which embeds target object priors with semantic modifications. To make the output corresponding to the target object, we utilize SDEdit~\cite{sdedit} to guide the image generation.
Inspired by \citet{darf}, the second introduces a monocular depth conditioning ControlNet~\cite{controlnet}, which is fine-tuned using data pair generation as in Sec.~\ref{Method: Gaussian Repair Model Setup}. We also assess the performance using masked depth conditioning.
Furthermore, we consider Zero123-XL~\cite{objaverseXL, zero123}, a well-known single-image reconstruction model requiring object-centered input images with precise camera rotations and positions. Here, we manually align the coordinate system and select the closest image to the novel viewpoint as its reference.

\begin{figure}[t]
\centering
\includegraphics[width=0.95\linewidth]{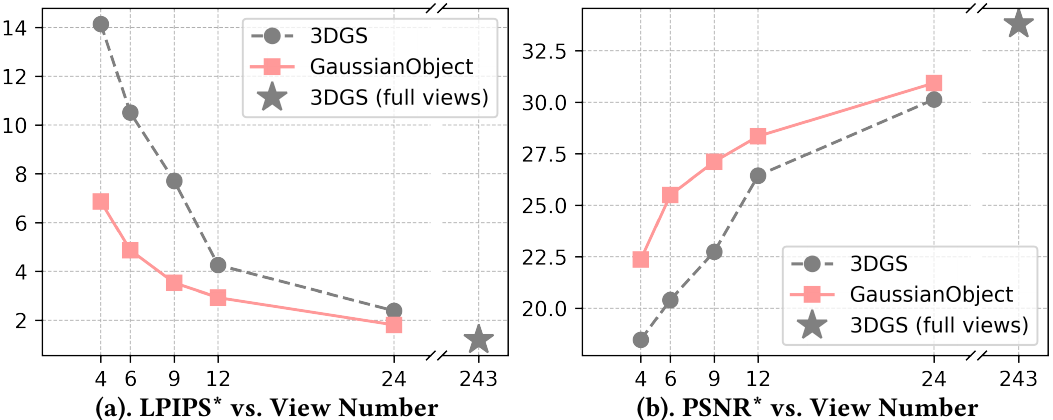}
\vspace{-1em}
\caption{Ablation on Training View Number. Experiments are conducted on scene \textit{kitchen} in the MipNeRF360 dataset.}
\Description{}
\label{fig: num_view}
\end{figure}

\begin{table}[t]
\newcommand{\xmark}{\ding{55}}
\centering
\caption{Ablation study about alternatives of the Gaussian repair model.}
\vspace{-1em}
\setlength{\tabcolsep}{5pt}
\label{tab: repair model structure}
\begin{tabular}{l|ccc}
\toprule
Method & \textbf{LPIPS$^*$} $\downarrow$ & \textbf{PSNR} $\uparrow$ & \textbf{SSIM} $\uparrow$ \\
\midrule
\rev{Zero123-XL~\cite{zero123}}       & \rev{13.97}     & \rev{17.71}     & \rev{0.8921} \\
Dreambooth~\cite{dreambooth}    & 6.58          & 21.85          & 0.9093          \\
Depth Condition & 7.00          & 21.87          & 0.9112          \\
Depth Condition w/ Mask  & 6.87          & 21.92          & 0.9117          \\
\midrule
\name (Ours)          & \textbf{5.79} & \textbf{23.55} & \textbf{0.9220} \\
\bottomrule
\end{tabular}
\vspace{-1em}
\end{table}

The results, as shown in Table~\ref{tab: repair model structure} and Fig.~\ref{fig: repair model structure}, reveal that semantic modifications proposed by Dreambooth alone fail in 3D-coherent synthesis. Monocular depth conditioning, whether with or without masks, despite some improvements, still struggles with depth roughness and artifacts. 
Zero123-XL, while generating visually acceptable images, the multi-view structure consistency is lacking. In contrast, our model excels in both 3D consistency and detail fidelity, outperforming others qualitatively and quantitatively.

\textit{Effect of View Numbers.}
We design experiments to evaluate the advantage of our method over different training views. As shown in Fig.~\ref{fig: num_view}, \name consistently outperforms vanilla 3DGS in varying numbers of training views. Besides, \name with 24 training views achieves performance comparable to that of 3DGS trained on all views (243).

\subsection{Limitations and Future Work}
\begin{figure}[t]
\centering
\includegraphics[width=0.85\linewidth]{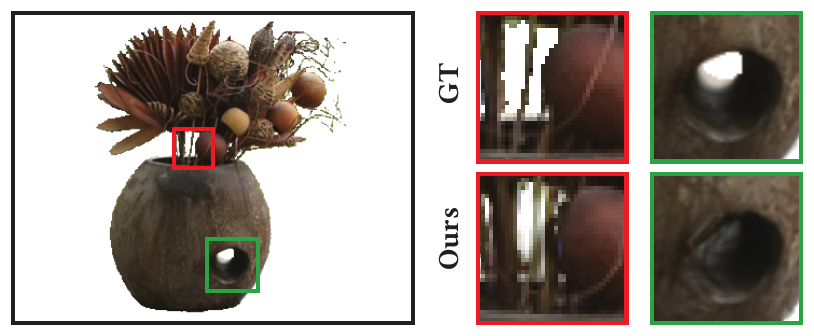}
\vspace{-1em}
\caption{Hallucinations of non-existent details. \name may fabricate visually reasonable details in areas with little information. For instance, the hole in the stone vase is filled in.}
\Description{}
\label{fig: limitation hallucinations}
\end{figure}

\begin{figure}[t]
\centering
\includegraphics[width=0.85\linewidth]{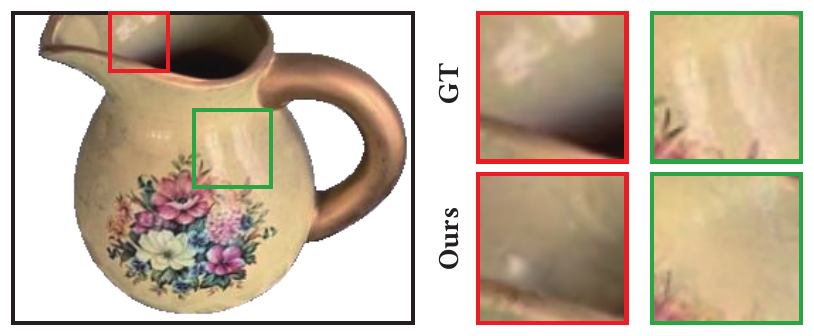}
\vspace{-1em}
\caption{Comparative visualization highlighting the challenge of reconstructing view-dependent appearance with only four input images.}
\Description{}
\label{fig: limitation view dependent}
\end{figure}

\name demonstrates notable performance in sparse 360\deg object reconstruction, yet several avenues for future research exist.
In regions completely unobserved or insufficiently observed, our repair model may generate hallucinations, \textit{i.e.}, it may produce non-existent details, as shown in Fig.~\ref{fig: limitation hallucinations}. However, these regions are inherently non-deterministic in information, and other methods also struggle in these areas. 
Additionally, due to the high sparsity level, our model is currently limited in capturing view-dependent effects. With such sparse data, our method cannot differentiate whether the appearance is view-dependent or inherent. 
Consequently, it `bakes in' the view-dependent features (like reflected white light) onto the surface, resulting in an inability to display view-dependent appearance from novel viewpoints correctly and leading to some unintended artifacts as demonstrated in Fig.~\ref{fig: limitation view dependent}.
Fine-tuning diffusion models with more view-dependent data may be a promising direction. \rev{Besides, integrating GaussianObject with surface reconstruction methods like 2DGS~\cite{2dgs} and GOF~\cite{gof} is a promising direction.}
Furthermore, CF-\name achieves competitive performance, but there is still a performance gap compared to precise camera parameters. An interesting exploration is to leverage confidence maps from matching methods for more accurate pose estimation.

\section{Conclusion}

In summary, \name is a novel framework designed for high-quality 3D object reconstruction from extremely sparse 360\deg views, based on 3DGS with real-time rendering capabilities. 
We design two main methods to achieve this goal: structure-prior-aided optimization for facilitating the multi-view consistency construction and a Gaussian repair model to remove artifacts caused by omitted or highly compressed object information. 
We also provide a COLMAP-free version that can be easily applied in real life with competitive performance.
We sincerely hope that \name can advance daily-life applications of reconstructing 3D objects, markedly reducing capture requirements and broadening application prospects.

\begin{acks}
This work was supported by the NSFC under Grant 62322604 and 62176159, and in part by the Shanghai Municipal Science and Technology Major Project under Grant 2021SHZDZX0102. The authors express gratitude to the anonymous reviewers for their valuable feedback and to Deyu Wang for his assistance with figure drawing and Blender support.
\end{acks}

{
\bibliographystyle{ACM-Reference-Format}
\bibliography{sample-base}
}

\newpage

\appendix

\section{Appendix}
The contents of this appendix include:
\begin{enumerate}
    \item Dataset Details (\ref{App: dataset datails}).
    \item \rev{Implementation} Details (\ref{App: Implement Details}).
    \item Experiment Details of Comparison (\ref{App: Experiment Details of Comparison}).
    \item More Experimental Results (\ref{App: More Experimental Results}).
    \item Ablation Study Details (\ref{App: Ablation Study Details}).
    \item Supplementary Video Descriptions (\ref{App: Supplementary Video Descriptions}).
\end{enumerate}

\subsection{Dataset Details} \label{App: dataset datails}
To ensure a rigorous evaluation, we employ SAM-based methods~\cite{sam,sa3d} to generate consistent object masks of test views. While this process is necessary for benchmarking in our study, it is not required for practical applications. In real-world scenarios, the system necessitates only four masks from captured images, which are straightforward to obtain with any segmentation method. 

\paragraph{MipNeRF360} The sparse MipNeRF360 dataset, derived from the dataset provided by \citet{mip360}, focuses on three scenes containing a primary object: bonsai, garden, and kitchen. For performance evaluation, the scenes are tested using images downsampled by a factor of 4$\times$, following the train-test splits from \citet{reconfusion}.

\paragraph{OmniObject3D} OmniObject3D includes 6k real 3D objects in 190 large-vocabulary categories. We selected 17 objects from OmniObject3D \cite{OmniObject3D}. The items chosen include: \textit{back-pack\_016}, \textit{box\_043}, \textit{broccoli\_003}, \textit{corn\_007}, \textit{dinosaur\_006}, \textit{flower\_pot\_007}, \textit{gloves\_009}, \textit{guitar\_002}, \textit{hamburger\_012}, \textit{picnic\_basket\_009}, \textit{pineapple\_013}, \textit{sandwich\_003}, \textit{suitcase\_006}, \textit{timer\_010}, \textit{toy\_plane\_005}, \textit{toy\_truck\_037}, and \textit{vase\_012} for a fair evaluation.
We manually choose camera views for training and use every eighth view left for testing. Most scenes are originally in 1080p resolution, while \textit{gloves\_009} and \textit{timer\_010} are in 720p resolution. To maintain consistency across all scenes, we upscaled the images from these two scenes to 1080p resolution. All the images are downsampled to a factor of 2$\times$.

\paragraph{OpenIllumination} OpenIllumination is a real-world dataset captured by the LightStage. We use the sparse OpenIllumination dataset proposed in ZeroRF~\cite{zerorf}. 
Note that ZeroRF re-scaled and re-centered the camera poses to align the target object with the world center.
We test on the sparse OpenIllumination dataset~\cite{OpenIllumination} with provided object masks, which are generated by SAM and train-test splits, downscaling the images by a factor of 4$\times$, following the same experimental setting as in~\citet{zerorf}.

\paragraph{Our-collected Unposed Images} 
To better align with daily usage, we captured four images of common objects using an iPhone 13 from approximately front, back, left, and right directions without strict requirements on camera positioning or angles. We employed DUSt3R~\cite{dust3r} to predict the camera parameters, including intrinsics and poses, for these images. Additionally, we used SAM~\cite{sam} along with the iPhone’s native segmentation features to perform image segmentation. All comparative methods also included these camera parameters and masks.

\begin{figure*}[!t]
\centering
\includegraphics[width=0.96\linewidth]{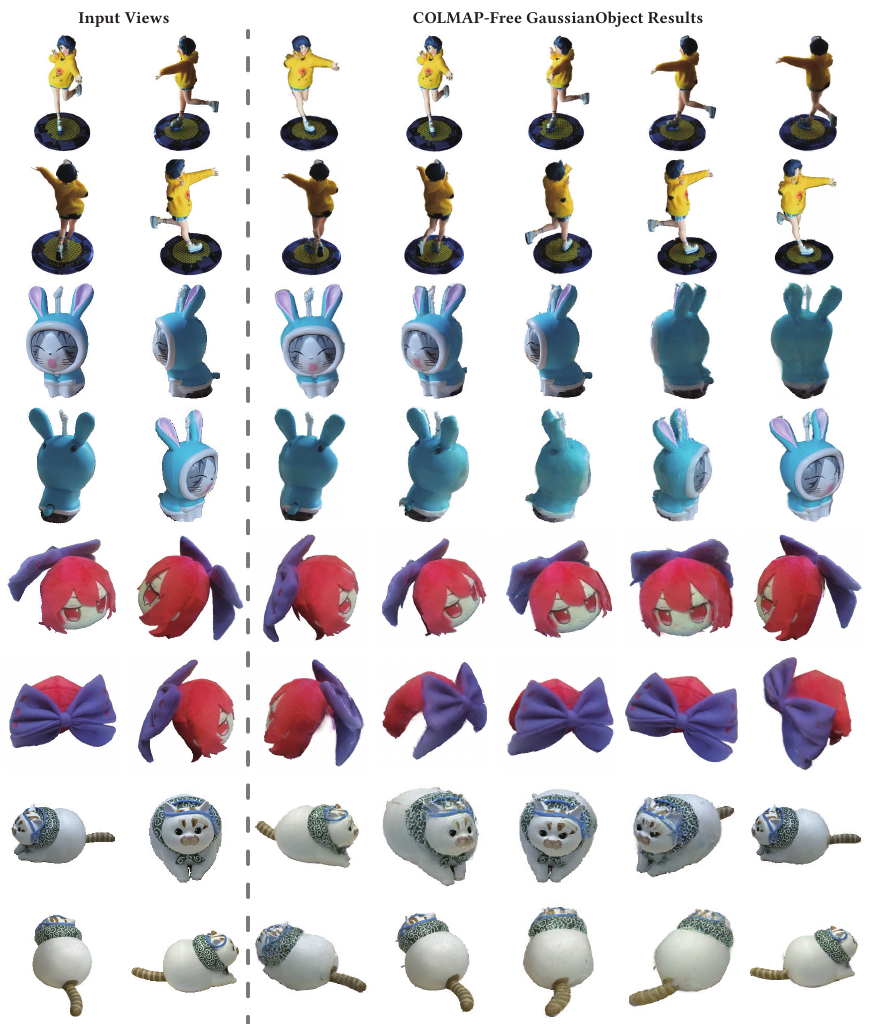}
\caption{Performance of CF-GaussianObject among iPhone captured images. By introducing modified DUSt3R into our GaussianObject, we successfully achieve COLMAP-free reconstruction from extremely sparse views.}
\Description{}
\label{fig: casual capture}
\end{figure*}

\subsection{\rev{Implementation} Details} \label{App: Implement Details}
We build our \name upon 3DGS~\cite{3dgs} and Threestudio~\cite{threestudio2023}. 
For visual hull reconstruction, we adopt a coarse-to-fine approach, starting with rough spatial sampling to estimate the bounding box and then proceeding to detailed sampling within it. Since the sparse input views cannot provide sufficient multi-view consistency, we reconstruct all objects with only two degrees of spherical harmonics (SH). 
\rev{We adopt the densification and opacity reset methods from vanilla 3DGS, conducting densification every 100 iterations and resetting opacity every 1,000 iterations.}
As for floater elimination, we start it from 500 iterations and conduct it every 500 iterations until 6k iterations. The adaptive threshold $\tau$ is initially set to the mean plus the standard deviation of the average distance calculated via KNN and is linearly decreased to 0 in 6k iterations. The loss weight for $\mathcal{L}_{\text{gs}}$ is set to: $\lambda_{\text{SSIM}} = 0.2$, $\lambda_{\text{m}} = 0.001$ and $\mathcal{L}_{\text{d}} = 0.0005$.

\begin{figure*}[htbp]
\centering
\includegraphics[width=0.6\linewidth]{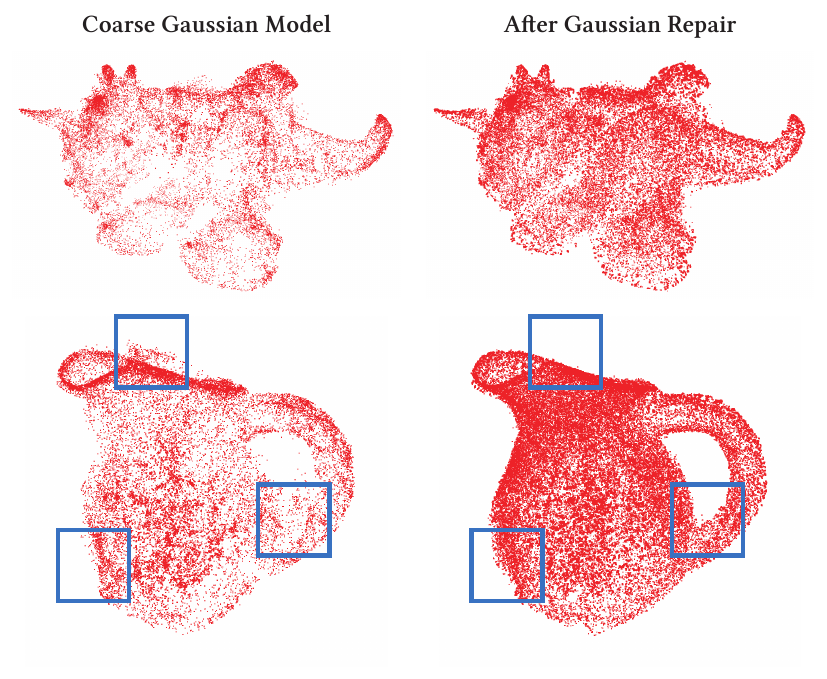}
\caption{Point clouds of \name before and after Gaussian repair. In the \textit{dinosaur} scene, the Gaussian repair model significantly enhances the point cloud density, resulting in a more distinct and clear representation of the dinosaur's body. In the \textit{vase} scene, noticeable artifacts around the handle and the mouth are effectively eliminated by the Gaussian repair model, and the incomplete bottom of the vase is repaired. These key areas are highlighted with blue boxes.}
\Description{}
\label{fig: structure repair}
\end{figure*}

For the Gaussian repair model setup, we use leave-one-out training to generate image pairs and 
3D Noise $\epsilon_{s}$. 
We use the visual hull corresponding to $N$ reference images to initialize all 3DGS models during leave-one-out training. We add noises to all the attributes of 3D Gaussians except for SH coefficients. Whenever we need a data pair from adding 3D noises, we use the mean $\mu_{\Delta}$ and variance $\sigma_{\Delta}$ to generate new noisy Gaussians for rendering, thus enabling sufficient data generation. All the images are constantly padded or center-cropped before being fed into the Gaussian repair model to preserve the real ratio of the target object. At the first iteration of training the Gaussian repair model, manual noise generated according to distribution is used for training with a $100\%$ probability. Each time training is conducted with manual noise, this probability is reduced by $0.5\%$, to utilize cached images from leave-one-out training increasingly\rev{, as illustrated in Algorithm~\ref{alg:gaussian repair}}. Referred to Dreambooth~\cite{dreambooth}, the [V] used for object-specific text prompt is ``xxy5syt00''. We apply LoRA~\cite{lora} in the fine-tuning process. LoRA optimizes a learnable low-rank residual matrix $\Delta W_i = A_iB_i$, where $A_i \in \mathbb{R}^{m \times r}$, $B_i \in \mathbb{R}^{r \times n}$. The hyperparameter of the LoRA rank $r \ll m, n$. We add LoRA layers to all embedding, linear, and convolution layers of the transformer blocks in the diffusion U-Net, the ControlNet U-Net, and the CLIP model. LoRA rank $r$ is set to $64$, and the learning rate is set to $10^{-3}$. We fine-tune the Gaussian repair model for 1800 steps, optimized using an AdamW optimizer~\cite{adamw} with $\beta$ values of $(0.9, 0.999)$.

\begin{algorithm}[htbp]
\caption{\rev{Gaussian Repair Model Data Generation Algorithm}}
\label{alg:gaussian repair}
\renewcommand{\algorithmicrequire}{\textbf{Input:}}
\renewcommand{\algorithmicensure}{\textbf{Output:}}
\begin{algorithmic}[1]
\REQUIRE Gaussian Repair Model $\mathcal{R}$, Coarse 3DGS Model $\mathcal{G}_c$, Mean $\mu_{\Delta,a}$ and Variance $\sigma_{\Delta,a}$ for Each Attribute $a$ of $\mathcal{G}_c$, $N$ Leave-one-out 3DGS Models $\{\hat{\mathcal{G}}_c^i\}_{i=0}^{N-1}$
\STATE $P_\text{manual}\gets1$
\STATE Sample $p$ from $U[0,1]$
\FOR{each iteration}
    \STATE Sample an index $i$ from $\{0,1,\dots,N-1\}$
    \IF{$p<P_\text{manual}$}
        \STATE $\mathcal{G}_c(\epsilon_s)\gets\mathcal{G}_c$
        \FORALL{attribute $a$ of $\mathcal{G}_c$ except for SH coefficients}
            \STATE Sample $\epsilon_{s,a}$ from $N(\mu_{\Delta,a},\sigma_{\Delta,a}^2)$
            \STATE $a\gets a+\epsilon_{s,a}$
        \ENDFOR
        \STATE  $x'\gets x'(\mathcal{G}_c(\epsilon_s),\pi^i)$
        \STATE $P_\text{manual}\gets0.995\times P_\text{manual}$
    \ELSE
        \STATE $x'\gets x'(\hat{\mathcal{G}}_c^i,\pi_i)$
    \ENDIF
    \STATE Optimize $\mathcal{R}$ with data pair $(x',x_i)$
\ENDFOR
\end{algorithmic}
\end{algorithm}

For the Gaussian repair process, we optimize the coarse 3DGS model for 4k steps, involving both a reference image and a repaired novel view image for the first 2800 steps supervised by $\mathcal{L}_{\text{rep}}$ and $\mathcal{L}_{\text{gs}}$ together, with the weight of $\mathcal{L}_{\text{rep}}$ progressively decayed from $1.0$ to $0.1$. The final 1200-step training only involves reference views.
Based on the camera poses of $N$ reference views, we estimate an elliptic trajectory that encircles the object and find $N$ points on the trajectory closest to the reference views, respectively. These $N$ points divide the ellipse into $N$ segments of the arc. On each arc segment, novel views are only sampled from the middle $80\%$ of the arc (repair path). The remaining arcs constitute reference paths. To expedite training, we avoid using the time-consuming DDIM process~\cite{ddim} for every novel view rendering. Instead, we employ cached images for optimization. Specifically, at the beginning of every 200 iterations, we sample and repair two novel views from each repair path. Throughout these 200 iterations, we utilize these images to repair 3D Gaussians.
We set the weights for $\mathcal{L}_{\text{rep}}$ as follows: $\lambda_1 = 0.5$, $\lambda_2 = 0.5$ and $\lambda_p = 2.0$ accross all experiments.
We employ densification and opacity resetting to regulate the number of Gaussians. The Gaussian densification process is effective from 400 to 3600 steps. The Gaussian model undergoes densification and pruning at intervals of every 100 steps, and its opacity is reset every 500 steps.

For CF-GaussianObject, we modify the DUSt3R~\cite{dust3r} pipeline to better suit our task. We assume that all cameras share the same intrinsic, while the original DUSt3R predicts different camera focal lengths for each view. We initially use the average camera focal estimated from input views and further optimize it using DUSt3R’s global optimization process. Background points are removed from DUSt3R’s point cloud $\mathcal{P}$ using object masks and the predicted camera poses via a back-projection algorithm. The resultant $\mathcal{P}$ is much sparser than the visual hull point cloud, missing points on unseen surfaces and object fillers. we augment $\mathcal{P}$ with $10\%$ of the points from the visual hull point cloud to densify the initial 3D Gaussian points. During the first 4000 steps of the 3DGS initial optimization process, we employ the tracking losses from MonoGS~\cite{monogs} and InstantSplat~\cite{instantsplat} to further optimize the camera poses $\hat{\Pi}^{\text{ref}}$ predicted by DUSt3R.
\begin{equation}
\hat{\pi}^{\text{ref}*}=\mathop{\arg\min}_{\hat{\pi}^\text{ref}}\|\mathcal{G}_c(\hat{\pi}^\text{ref})-x^\text{ref}\|_1+\lambda_\text{pose}\|\hat{\pi}^\text{ref}-\hat{\pi}^\text{ref}_0\|_1,
\end{equation}
where $\hat{\pi}^{\text{ref}*}$ denotes the refined camera pose, and $\hat{\pi}^\text{ref}_0$ represents the initial camera pose from DUSt3R. $\lambda_\text{pose}$ is introduced to ensure the refined poses do not deviate excessively from the initial ones, which is set to $0.0005$ in our experiments. To evaluate CF-GaussianObject, we align the SfM camera poses with the noisy ones and optimize each test camera for up to 400 steps with the same loss during test time.

\subsection{Experiment Details of Comparison} \label{App: Experiment Details of Comparison}

FSGS~\cite{fsgs} requires SfM points generated from the input images, which are too sparse in our setting of 4 input views. Alternatively, we randomly pick $N_{\mathrm{pick}}$ points from the SfM points with full images as input:
\begin{equation}
N_{\mathrm{pick}}=\max\left(150, \frac{k_{\mathrm{sparse}}}{k_{\mathrm{all}}}N_{\mathrm{all}}\right),
\end{equation}
where $k_{\mathrm{sparse}}$ is the number of input images, $k_{\mathrm{all}}$ is the total number of images, and $N_{\mathrm{all}}$ is the total number of points.

\begin{figure}[t]
\centering
\includegraphics[width=\linewidth]{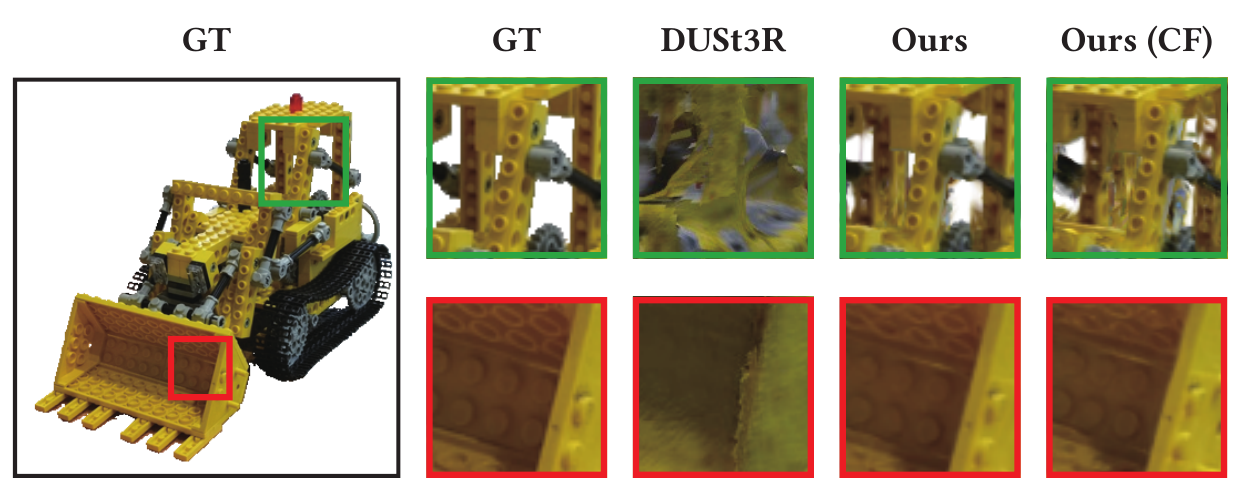}
\caption{Qualitative comparisons between DUSt3R and CF-GaussianObject with four input views.}
\Description{}
\label{fig: dust3r}
\end{figure}

\subsection{More Experimental Results} \label{App: More Experimental Results}
\subsubsection{Per-scene Performance}
\begin{figure*}[thbp]
\centering
\includegraphics[width=0.9\linewidth]{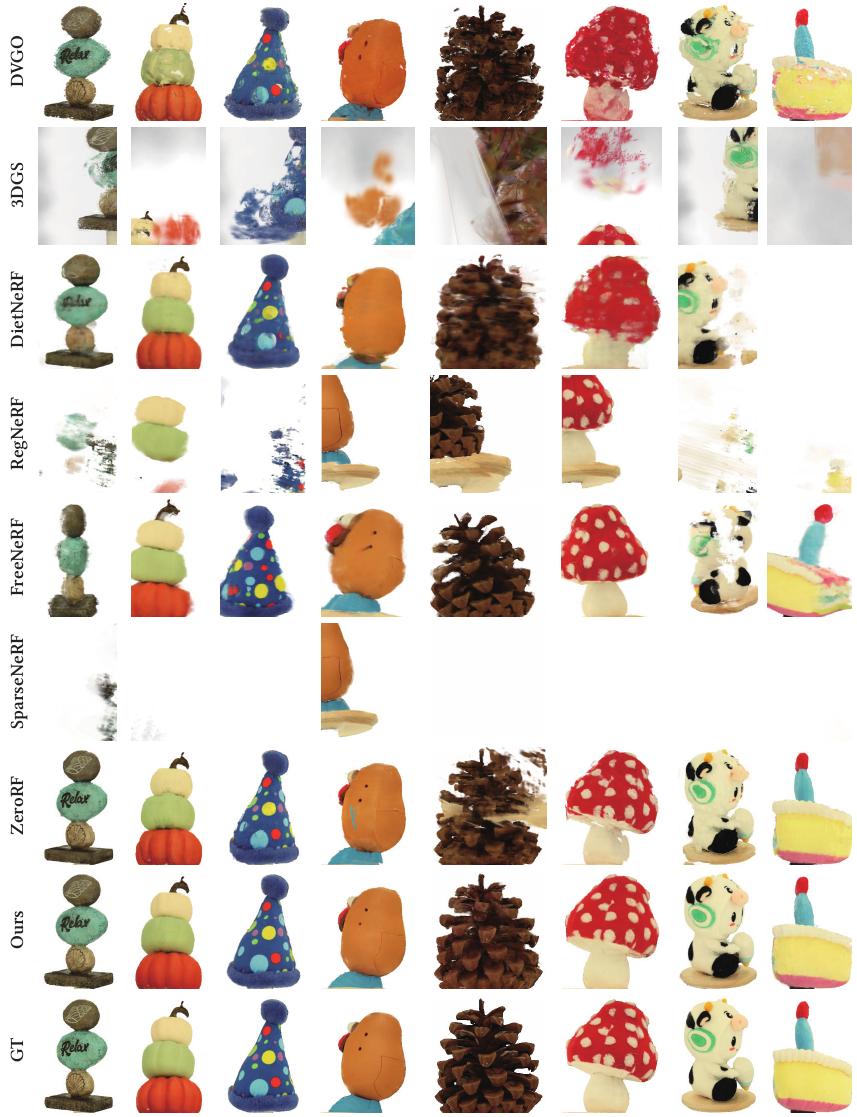}
\caption{Qualitative examples on the OpenIllumination dataset with four input views.}
\label{fig: qualitative results oppo}
\end{figure*}

We provide per-scene qualitative metrics of MipNeRF360 in Tab.~\ref{tab: per scene mip360 4}, Tab.~\ref{tab: per scene mip360 6} and Tab.~\ref{tab: per scene mip360 9}; of OmniObject3D in Tab.~\ref{tab: per scene omni3d 4.1}, Tab.~\ref{tab: per scene omni3d 4.2}, Tab.~\ref{tab: per scene omni3d 6.1}, Tab.~\ref{tab: per scene omni3d 6.2}, Tab.~\ref{tab: per scene omni3d 9.1} and Tab.~\ref{tab: per scene omni3d 9.2}; of OpenIllumination in Tab.~\ref{tab: per scene oppo 4} and Tab.~\ref{tab: per scene oppo 6}. Additional qualitative comparisons on the OpenIllumination dataset are shown in Fig.~\ref{fig: qualitative results oppo}.

\subsubsection{Performance of CF-GaussianObject}
Our COLMAP-free CF-GaussianObject bypasses the traditional Structure-from-Motion (SfM) pipeline requirements, enabling application to casually captured images. Utilizing an iPhone 13, we capture four images of common objects and obtain their corresponding masks using SAM~\cite{sam}. We then employ CF-GaussianObject for the reconstruction of these images, with the results of the novel view synthesis displayed in Fig.~\ref{fig: casual capture}. CF-GaussianObject delivers high-quality reconstructions from just four images without the need for precise camera parameters, producing images with exceptional visual quality and detailed richness. These results underscore CF-GaussianObject's effectiveness in achieving high-fidelity reconstruction from sparsely captured real-world images, showcasing its substantial potential for practical applications.

\subsubsection{Comparison with DUSt3R}
To further show the effectiveness of our design, we present qualitative comparisons between CF-\name and DUSt3R~\cite{dust3r} in Fig.~\ref{fig: dust3r}. These results confirm that the strong performance stems from our design. Overall, these findings highlight the generalizability of CF-\name and its potential for real-world applications.

\begin{figure}[t]
\centering
\includegraphics[width=\linewidth]{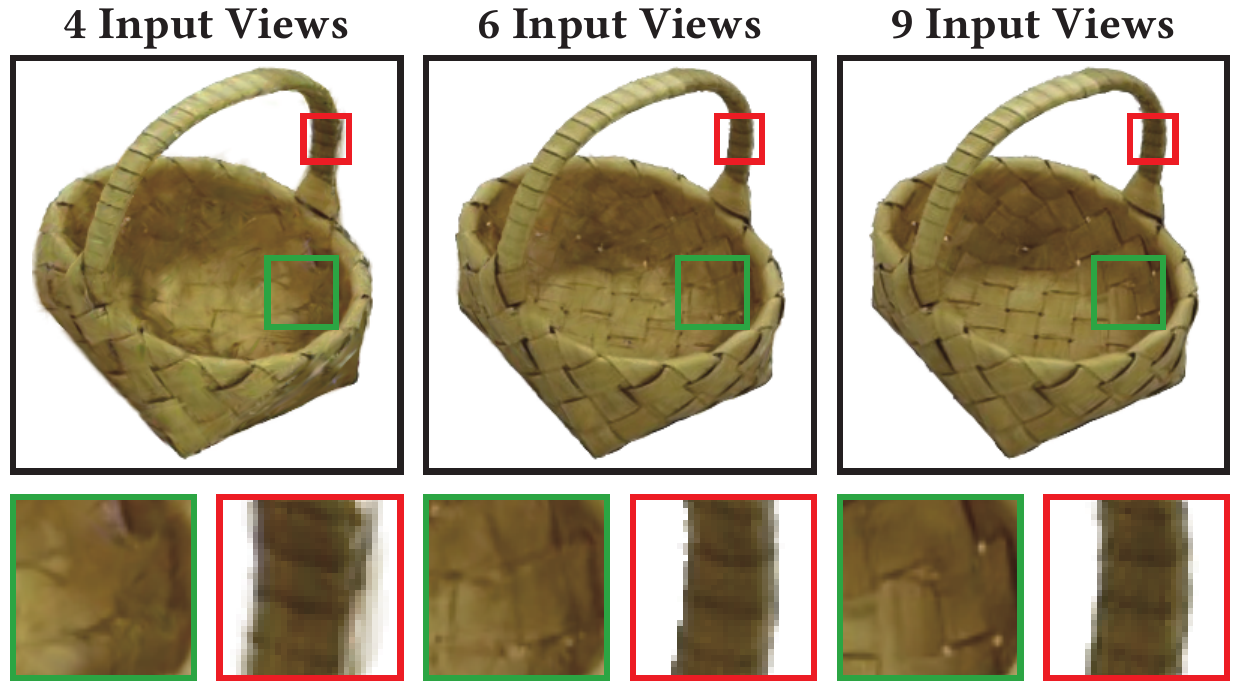}
\vspace{-1em}
\caption{Rendering performance with varying training views.}
\Description{}
\label{fig: 469}
\end{figure}

\subsubsection{Influence of Gaussian repair process}
\begin{figure*}[thbp]
\centering
\includegraphics[width=0.95\linewidth]{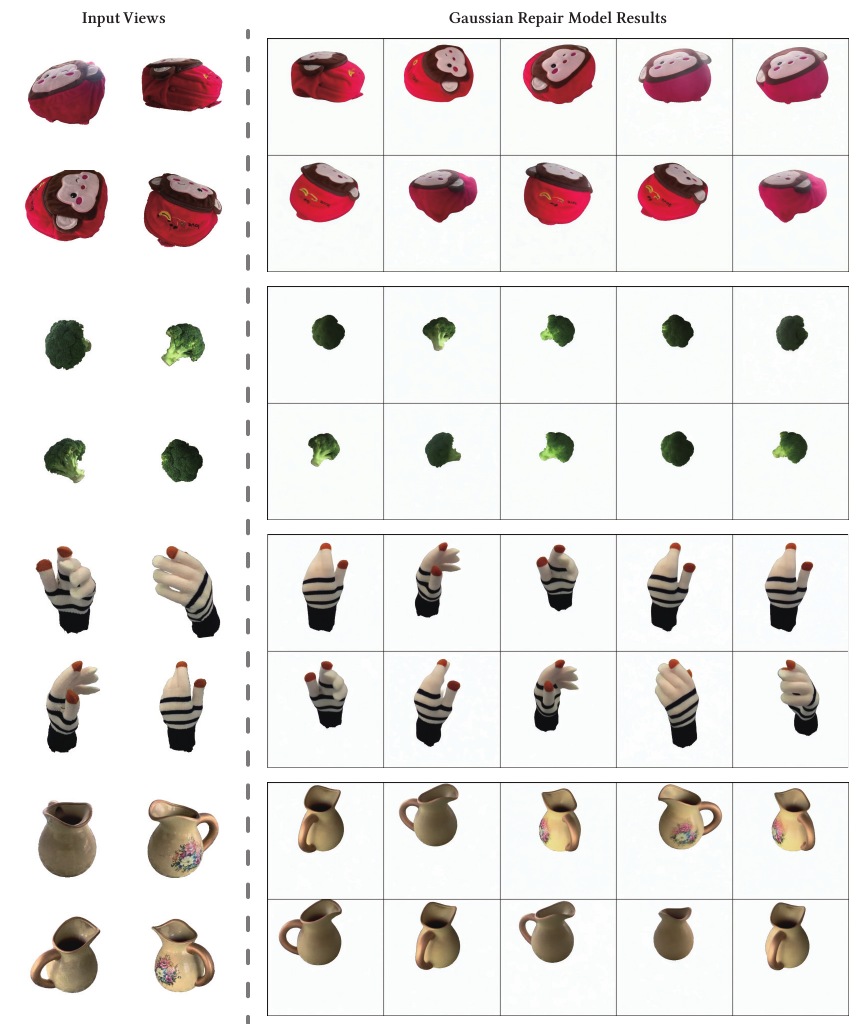}
\caption{Qualitative samples of the Gaussian repaired models on several scenes from different views.}
\label{fig: qualitative results repair}
\end{figure*}

We provide extra validations on our Gaussian repair process. Figure~\ref{fig: structure repair} illustrates the distribution of coarse 3D Gaussians and the refined distribution after the Gaussian repair. This comparison clearly demonstrates the significant enhancement in geometry achieved by our repair process. We also provide qualitative samples of the Gaussian repair model in Fig.~\ref{fig: qualitative results repair}, showing that the Gaussian repair model can effectively generate high-quality and consistent images from multiple views.

\begin{table}[th]
\centering
\caption{Comparison of training time on one single RTX 3090 GPU.}
\begin{tabular}{lc}
\toprule
Method        & Time    \\
\midrule
DietNeRF~\cite{dietnerf}       & 8h+    \\
RegNeRF~\cite{regnerf}        & 48h+   \\
FreeNeRF~\cite{freenerf}       & 24h+   \\
SparseNeRF~\cite{sparsenerf}     & 24h+   \\
ZeroRF~\cite{zerorf}         & $\sim$35min \\
ReconFusion~\cite{reconfusion}    & 8h+    \\
GaussianObject & $\sim$30min \\
CF-GaussianObject (ours) & \rev{$\sim$33min} \\
\bottomrule
\end{tabular}
\label{tab: runtime_comparison}
\end{table}

\subsubsection{Training Time Comparison}
We show the training time of GaussianObject and the various baselines on one single RTX 3090 GPU in Table~\ref{tab: runtime_comparison}. 
GaussianObject can be finished around 30 minutes, much faster than previous methods with higher quality. 
\rev{The process breakdown for GaussianObject is as follows: Initial optimization with structure priors takes approximately 1 minute; Gaussian repair model setup requires about 15 minutes; and Gaussian repair with distance-aware sampling lasts around 14 minutes.}

\rev{For the COLMAP-free version, we incorporate a customized diff-gaussian-rasterization module during the coarse GS optimization. This module, which is not used in the standard GaussianObject nor the repair model setup and Gaussian repair, necessitates additional computations for gradient descent on camera poses and uses a faster rasterization module elsewhere. The training time for CF-GaussianObject typically totals about 33 minutes, broken down as follows: DUSt3R~\cite{dust3r} completes in roughly 17 seconds; initial optimization with structure priors takes about 3 minutes due to the customized diff-gaussian-rasterization module; Gaussian repair model setup remains at 15 minutes; and Gaussian repair with distance-aware sampling also takes about 14 minutes.}

\begin{table}[t]
\newcommand{\xmark}{\ding{55}}
\centering
\caption{Robustness among randomness on MipNeRF360 dataset.}
\setlength{\tabcolsep}{5pt}
\label{tab: random seed}
\begin{tabular}{l|ccc}
\toprule
Performance & \textbf{LPIPS$^*$} $\downarrow$ & \textbf{PSNR} $\uparrow$ & \textbf{SSIM} $\uparrow$ \\
\midrule
Mean         & 4.97         & 24.82          & 0.9345          \\
Std          & 0.06         & 0.169          & 0.0007          \\
\bottomrule
\end{tabular}
\end{table}

\subsubsection{Robustness of Random Noise}
Diffusion models rely on adding random noise followed by denoising to generate images; hence, their performance is heavily dependent on the random seed used. GaussianObject also employs a diffusion model to refine 3D Gaussian, so the diversity inherent in diffusion could affect the reconstruction quality of GaussianObject. To mitigate this diversity, we finetune the diffusion model and devise a distance-aware sampling strategy for Gaussian repair. To validate this, we conduct 10 independent experiments with different random seeds, and the results are presented in Table~\ref{tab: random seed}. Experimental results confirm the effectiveness of our design, demonstrating that randomness has little influence on our performance. 

\begin{figure*}[thbp]
\centering
\includegraphics[width=\linewidth]{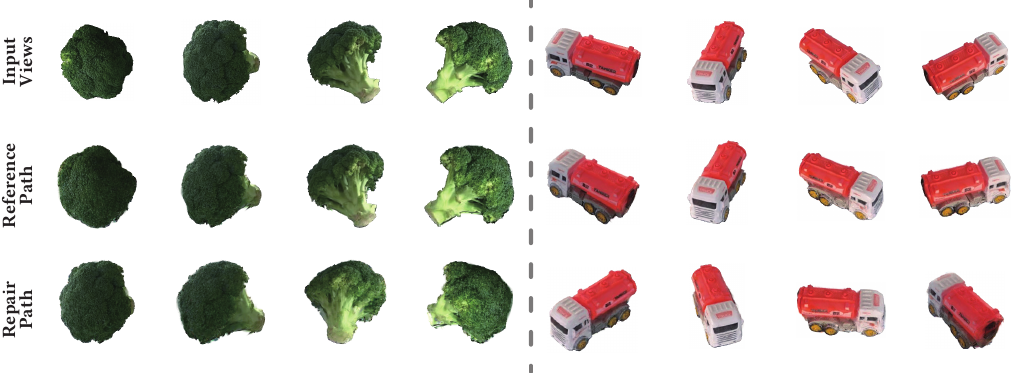}
\caption{Qualitative results from reference and repair path. The first row displays the four reference views, the second row shows the renderings from the reference path and the third row shows the renderings from the repair path. \name achieves high-quality renderings on both reference and repair paths, indicating the effectiveness of our Gaussian repair process.}
\Description{}
\label{fig: novel_view}
\end{figure*}

\subsubsection{Performance on Reference and Repair Path}
The rendering capabilities of GaussianObject are evaluated on both the reference and repair paths. To ensure the utmost fidelity to the reference views, the Gaussian repair process is exclusively applied along the repair path. Fig.~\ref{fig: novel_view} illustrates that GaussianObject delivers exceptional visual quality across both the reference and repair paths. Such results highlight the efficacy of our Gaussian repair process. \name consistently produces high fidelity and visual excellence renderings, irrespective of whether the views are from regions with sufficient or insufficient coverage.

\subsubsection{Performance with Varying Views}
We demonstrate the performance of GaussianObject under various training views in Fig.~\ref{fig: 469}. As more views are provided, GaussianObject can reconstruct more details. This demonstrates the model's capacity to leverage additional perspectives for improved accuracy and detail in the reconstructed output. Specifically, with an increasing number of views, the visual quality of the renderings was significantly enhanced, reducing artifacts and producing finer details. This improvement highlights the robustness and scalability of GaussianObject in handling complex visual data.

\subsubsection{\rev{Performance with view distribution}}

In this study, we evaluate the performance of GaussianObject on the `kitchen' scene from the MipNeRF360 dataset across various view distributions. We systematically report the angles between input views in the `view distribution' column of Table~\ref{tab: view distributions}, with the first row reflecting the configuration used in the manuscript. Results reveal that GaussianObject consistently achieves robust performance across diverse view distributions, a capability that eludes many current Large Reconstruction Models (LRMs)~\rev{\cite{lgm, gslrm2024, grm}}. This robustness is crucial for practical applications where varied viewpoints are common.

\begin{table}[ht]
\newcommand{\xmark}{\ding{55}}
\centering
\caption{\rev{Performance in terms of view distribution.}}
\setlength{\tabcolsep}{5pt}
\label{tab: view distributions}
\begin{tabular}{c|ccc}
\toprule
View Distribution & \textbf{LPIPS$^*$} $\downarrow$ & \textbf{PSNR} $\uparrow$ & \textbf{SSIM} $\uparrow$ \\
\midrule
110.4$^\circ$, 112.4$^\circ$, 65.1$^\circ$, 72.1$^\circ$ & 6.8716 & 22.36 & 0.9104 \\
~87.3$^\circ$, ~91.5$^\circ$, 99.4$^\circ$, 81.8$^\circ$ & 6.8279 & 22.63 & 0.9098 \\
~60.4$^\circ$, 164.2$^\circ$, 51.1$^\circ$, 84.3$^\circ$ & 7.3637 & 21.72 & 0.9039 \\
121.6$^\circ$, 121.8$^\circ$, 59.5$^\circ$, 57.1$^\circ$ & 6.9842 & 22.97 & 0.9094 \\
\bottomrule
\end{tabular}
\end{table}

\subsection{Ablation Study Details} \label{App: Ablation Study Details}
We employ ControlNet-Tile~\cite{controlnet_v11, controlnet} as the Gaussian repair model of \name. ControlNet-Tile, based on Stable Diffusion v1.5, is designed to repair low-resolution or low-fidelity images by taking them as conditions. In our ablation study, we use other diffusion models as our Gaussian repair model in Sec.4.4. First, we use Stable Diffusion XL Image-to-Image~\cite{sdedit}, which cannot directly take images as input conditions. Therefore, we add $50\%$ Gaussian noise to the novel view images and use the model to generate high-fidelity images by denoising the noisy images. Additionally, we evaluate ControlNet-ZoeDepth based on Stable Diffusion XL. ControlNet-ZoeDepth, which takes the ZoeDepth~\cite{zoedepth} as a condition, is used to generate images aligned with the depth map. However, ZoeDepth may incorrectly assume that a single object with a white background is on a white table, leading to erroneous depth map predictions. In such cases, we test two settings: one using the entire predicted depth map and another using the object-masked depth map. Due to the ambiguity of monocular predicted depth, we add partial noise on the input image to preserve the image information, similar to SDEdit. Specifically, we add $50\%$ Gaussian noise to the novel view images and set the weight scale of the ControlNet model to $0.55$. In this experimental setup, we obtained the results of the ablation experiments on the structure of the Gaussian repair model.

\subsection{Supplementary Video Descriptions} \label{App: Supplementary Video Descriptions}
We prepare four sets of video comparisons for evaluation. The first set, \textit{videos\_1}, features videos rendered along trajectories encircling several objects, showcasing the performance of all tested models. These videos are organized by scene and the number of input views and are placed in different directories for ease of analysis.

In the second video set, \textit{videos\_2}, we provide more experiment videos. Each video is edited to demonstrate our results clearly. The video titled \textit{refresh.mp4} compares all the tested models with \name on 9 scenes with 4 input views. Videos labeled with the suffix \textit{\_compare.mp4} focus on comparing individual methods with \name. The video titled \textit{transfer\_3DGS\_to\_\name.mp4} sequentially illustrates the enhancements we have implemented on the 3DGS, detailing each step of our improvement process. Additionally, the video \textit{COLMAP\_free.mp4} shows our COLMAP-free experiments using our own collected unposed images, presenting both the input images and reconstruction results, along with comparisons to other methods.

In the third video set, \textit{videos\_3}, we present videos from COLMAP-free experiments. Each video is labeled according to the dataset, scene, and number of input views involved in the experiment.

In the last video set, \textit{videos\_4}, we provide input images and results from COLMAP-free experiments using our own collected unposed images. A subfolder named \textit{other\_methods} includes videos of reconstruction results obtained by other methods.

\begin{table*}[thbp]
\newcommand{\xmark}{\ding{55}}
\centering
\caption{Comparisons of per-scene metrics of MipNeRF360 with 4 input views.}
\label{tab: per scene mip360 4}
\begin{tabular}{l|ccc|ccc|ccc}
\toprule
\multirow{2}{*}{Method} & \multicolumn{3}{c|}{bonsai} & \multicolumn{3}{c|}{garden} & \multicolumn{3}{c}{kitchen} \\
 & LPIPS $\downarrow$ & PSNR $\uparrow$ & SSIM $\uparrow$ & LPIPS $\downarrow$ & PSNR $\uparrow$ & SSIM $\uparrow$ & LPIPS $\downarrow$ & PSNR $\uparrow$ & SSIM $\uparrow$ \\ 
\midrule
DVGO~\cite{dvgo}             & 0.3324                     & 10.45                     & 0.6980                     & 0.0559                     & 21.75                     & 0.9652                     & 0.3447                     & 10.97                     & 0.7104                     \\
3DGS~\cite{3dgs}             & 0.1408                     & \cellcolor{metric-3}16.42 & 0.8458                     & 0.0417                     & \cellcolor{metric-3}26.05 & 0.9769                     & \cellcolor{metric-3}0.1414 & \cellcolor{metric-3}18.47 & \cellcolor{metric-3}0.8746 \\
DietNeRF~\cite{dietnerf}     & \cellcolor{metric-3}0.1333 & 16.30                     & \cellcolor{metric-3}0.8682 & \cellcolor{metric-2}0.0231 & \cellcolor{metric-2}28.63 & \cellcolor{metric-3}0.9780 & 0.1787                     & 11.77                     & 0.8453                     \\
RegNeRF~\cite{regnerf}       & 0.2736                     & 9.99                      & 0.7847                     & 0.0300                     & 20.89                     & \cellcolor{metric-2}0.9794 & 0.3094                     & 9.89                      & 0.7788                     \\
FreeNeRF~\cite{freenerf}     & 0.2172                     & 10.16                     & 0.8052                     & 0.0300                     & 20.89                     & 0.9760                     & 0.2578                     & 10.08                     & 0.7789                     \\
\bottomrule
\end{tabular}
\end{table*}

\begin{table*}[thbp]
\newcommand{\xmark}{\ding{55}}
\centering
\caption{Comparisons of per-scene metrics of MipNeRF360 with 6 input views.}
\label{tab: per scene mip360 6}
\begin{tabular}{l|ccc|ccc|ccc}
\toprule
\multirow{2}{*}{Method} & \multicolumn{3}{c|}{bonsai} & \multicolumn{3}{c|}{garden} & \multicolumn{3}{c}{kitchen} \\
 & LPIPS $\downarrow$ & PSNR $\uparrow$ & SSIM $\uparrow$ & LPIPS $\downarrow$ & PSNR $\uparrow$ & SSIM $\uparrow$ & LPIPS $\downarrow$ & PSNR $\uparrow$ & SSIM $\uparrow$ \\ 
\midrule
DVGO~\cite{dvgo}             & 0.3908                     & 10.11                     & 0.6531                     & 0.0416                     & 21.90                     & 0.9728                     & 0.3677                     & 10.88                     & 0.6769                     \\
3DGS~\cite{3dgs}             & 0.1133                     & 18.42                     & 0.8736                     & 0.0330                     & \cellcolor{metric-3}27.54 & \cellcolor{metric-3}0.9809 & 0.1051                     & 20.40                     & 0.8858                     \\
DietNeRF~\cite{dietnerf}     & \cellcolor{metric-2}0.0863 & \cellcolor{metric-2}22.39 & \cellcolor{metric-2}0.9085 & \cellcolor{metric-3}0.0274 & 20.89                     & 0.9760                     & 0.0951                     & 22.80                     & 0.9012                     \\
RegNeRF~\cite{regnerf}       & 0.2736                     & 9.34                      & 0.7736                     & 0.0300                     & 20.88                     & 0.9794                     & 0.3180                     & 10.00                     & 0.7726                     \\
FreeNeRF~\cite{freenerf}     & \cellcolor{metric-3}0.0904 & \cellcolor{metric-3}22.09 & \cellcolor{metric-3}0.9083 & 0.0300                     & 20.89                     & 0.9760                     & \cellcolor{metric-2}0.0847 & \cellcolor{metric-2}23.78 & \cellcolor{metric-2}0.9153 \\
\bottomrule
\end{tabular}
\end{table*}

\begin{table*}[thbp]
\newcommand{\xmark}{\ding{55}}
\centering
\caption{Comparisons of per-scene metrics of MipNeRF360 with 9 input views.}
\label{tab: per scene mip360 9}
\begin{tabular}{l|ccc|ccc|ccc}
\toprule
\multirow{2}{*}{Method} & \multicolumn{3}{c|}{bonsai} & \multicolumn{3}{c|}{garden} & \multicolumn{3}{c}{kitchen} \\
 & LPIPS $\downarrow$ & PSNR $\uparrow$ & SSIM $\uparrow$ & LPIPS $\downarrow$ & PSNR $\uparrow$ & SSIM $\uparrow$ & LPIPS $\downarrow$ & PSNR $\uparrow$ & SSIM $\uparrow$ \\ 
\midrule
DVGO~\cite{dvgo}             & 0.3658                     & 11.17                     & 0.6938                     & 0.0367                     & 22.09                     & 0.9747                     & 0.3672                     & 10.95                     & 0.6843                     \\
3DGS~\cite{3dgs}             & 0.0932                     & 20.68                     & 0.9004                     & 0.0222                     & 29.46                     & 0.9839                     & 0.0771                     & 22.74                     & 0.9149                     \\
DietNeRF~\cite{dietnerf}     & \cellcolor{metric-3}0.0706 & \cellcolor{metric-3}24.45 & \cellcolor{metric-3}0.9263 & 0.0274                     & 20.89                     & 0.9760                     & 0.0774                     & 25.30                     & \cellcolor{metric-3}0.9247 \\
RegNeRF~\cite{regnerf}       & 0.2875                     & 9.45                      & 0.7706                     & 0.0300                     & 20.89                     & 0.9794                     & 0.2735                     & 10.70                     & 0.8050                     \\
FreeNeRF~\cite{freenerf}     & 0.0788                     & 23.99                     & \cellcolor{metric-3}0.9263 & \cellcolor{metric-2}0.0164 & \cellcolor{metric-2}33.05 & \cellcolor{metric-2}0.9859 & \cellcolor{metric-2}0.0702 & \cellcolor{metric-2}25.96 & \cellcolor{metric-2}0.9335 \\
\bottomrule
\end{tabular}
\end{table*}

\begin{table*}[thbp]
\newcommand{\xmark}{\ding{55}}
\centering
\caption{Comparisons of per-scene metrics of OmniObject3D with 4 input views.}
\label{tab: per scene omni3d 4.1}
\begin{tabular}{l|ccc|ccc|ccc}
\toprule
\multirow{2}{*}{Method} & \multicolumn{3}{c|}{backpack\_016} & \multicolumn{3}{c|}{box\_043} & \multicolumn{3}{c}{broccoli\_003} \\
& LPIPS $\downarrow$ & PSNR $\uparrow$ & SSIM $\uparrow$ & LPIPS $\downarrow$ & PSNR $\uparrow$ & SSIM $\uparrow$ & LPIPS $\downarrow$ & PSNR $\uparrow$ & SSIM $\uparrow$ \\
\midrule
DVGO       & 0.2674                     & 11.41                     & 0.7794                     & 0.0814                     & 20.21                     & 0.9455                     & 0.0953                     & 16.14                     & 0.9229                     \\
3DGS       & 0.1542                     & 15.13                     & 0.8505                     & \cellcolor{metric-3}0.0536 & 18.57                     & \cellcolor{metric-3}0.9665 & 0.0677                     & 15.68                     & 0.9458                     \\
DietNeRF   & 0.2903                     & 11.30                     & 0.8183                     & 0.1866                     & 15.54                     & 0.8867                     & 0.1028                     & 14.65                     & 0.9172                     \\
RegNeRF    & 0.2987                     & 9.56                      & 0.8022                     & 0.1298                     & 16.40                     & 0.9455                     & 0.1109                     & 14.08                     & 0.9239                     \\
FreeNeRF   & 0.1358                     & 11.57                     & 0.8899                     & 0.0775                     & 17.59                     & 0.9516                     & 0.0570                     & 16.36                     & 0.9559                     \\
SparseNeRF & 0.2570                     & 9.78                      & 0.7985                     & 0.0960                     & 16.52                     & 0.9437                     & 0.1068                     & 14.18                     & 0.9208                     \\
\bottomrule
\end{tabular}
\end{table*}

\begin{table*}[thbp]
\newcommand{\xmark}{\ding{55}}
\centering
\caption{Comparisons of per-scene metrics of OmniObject3D with 4 input views. \textit{continued}}
\label{tab: per scene omni3d 4.2}
\begin{tabular}{l|ccc|ccc|ccc|ccc}
\toprule
\multirow{2}{*}{Method} & \multicolumn{3}{c|}{picnic\_basket\_009} & \multicolumn{3}{c|}{pineapple\_013} & \multicolumn{3}{c|}{sandwich\_003} & \multicolumn{3}{c}{suitcase\_006} \\
& LPIPS $\downarrow$ & PSNR $\uparrow$ & SSIM $\uparrow$ & LPIPS $\downarrow$ & PSNR $\uparrow$ & SSIM $\uparrow$ & LPIPS $\downarrow$ & PSNR $\uparrow$ & SSIM $\uparrow$ & LPIPS $\downarrow$ & PSNR $\uparrow$ & SSIM $\uparrow$ \\
\midrule
DVGO       & 0.1540                     & 16.08                     & 0.8887                     & 0.1677                     & 14.39                     & 0.8747                     & 0.0423                     & 25.15                     & 0.9787                     & 0.1865                     & 14.52                     & 0.8697                     \\
3DGS       & 0.0837                     & 19.60                     & 0.9167                     & \cellcolor{metric-3}0.0573 & 16.74                     & \cellcolor{metric-2}0.9484 & 0.0285                     & 22.44                     & 0.9850                     & 0.1028                     & 15.18                     & 0.9331                     \\
DietNeRF   & 0.1512                     & 15.02                     & 0.8940                     & 0.0933                     & 16.48                     & 0.9336                     & 0.0253                     & 22.52                     & 0.9851                     & \cellcolor{metric-3}0.0542 & \cellcolor{metric-2}26.60 & \cellcolor{metric-2}0.9639 \\
RegNeRF    & 0.1313                     & 14.73                     & 0.9303                     & 0.1431                     & 14.01                     & 0.9005                     & 0.0293                     & 22.52                     & 0.9875                     & 0.2636                     & 12.03                     & 0.8587                     \\
\bottomrule
\end{tabular}
\end{table*}

\begin{table*}[thbp]
\newcommand{\xmark}{\ding{55}}
\centering
\caption{Comparisons of per-scene metrics of OmniObject3D with 6 input views.}
\label{tab: per scene omni3d 6.1}
\begin{tabular}{l|ccc|ccc|ccc}
\toprule
\multirow{2}{*}{Method} & \multicolumn{3}{c|}{backpack\_016} & \multicolumn{3}{c|}{box\_043} & \multicolumn{3}{c}{broccoli\_003} \\
& LPIPS $\downarrow$ & PSNR $\uparrow$ & SSIM $\uparrow$ & LPIPS $\downarrow$ & PSNR $\uparrow$ & SSIM $\uparrow$ & LPIPS $\downarrow$ & PSNR $\uparrow$ & SSIM $\uparrow$ \\
\midrule
DVGO       & 0.2363                     & 12.56                     & 0.8040                     & 0.1028                     & 19.74                     & 0.9412                     & 0.0853                     & 18.16                     & 0.9385                     \\
3DGS       & 0.0901                     & 21.26                     & 0.9122                     & 0.0582                     & 18.16                     & 0.9607                     & \cellcolor{metric-3}0.0483 & 16.77                     & 0.9621                     \\
DietNeRF   & 0.2117                     & 11.25                     & 0.8546                     & 0.1111                     & 17.25                     & 0.9325                     & 0.0994                     & 15.03                     & 0.9267                     \\
RegNeRF    & 0.2818                     & 10.08                     & 0.8291                     & 0.1251                     & 15.85                     & 0.9420                     & \cellcolor{metric-3}0.0483 & 16.84                     & \cellcolor{metric-3}0.9663 \\
FreeNeRF   & 0.2255                     & 12.30                     & 0.8181                     & 0.0530                     & 18.68                     & 0.9656                     & \cellcolor{metric-3}0.0483 & 16.84                     & 0.9625                     \\
SparseNeRF & 0.4828                     & 11.35                     & 0.7691                     & 0.1030                     & 18.16                     & 0.9602                     & \cellcolor{metric-3}0.0483 & 16.84                     & 0.9625                     \\
\bottomrule
\end{tabular}
\end{table*}

\begin{table*}[thbp]
\newcommand{\xmark}{\ding{55}}
\centering
\caption{Comparisons of per-scene metrics of OmniObject3D with 6 input views. \textit{continued}}
\label{tab: per scene omni3d 6.2}
\begin{tabular}{l|ccc|ccc|ccc|ccc}
\toprule
\multirow{2}{*}{Method} & \multicolumn{3}{c|}{picnic\_basket\_009} & \multicolumn{3}{c|}{pineapple\_013} & \multicolumn{3}{c|}{sandwich\_003} & \multicolumn{3}{c}{suitcase\_006} \\
& LPIPS $\downarrow$ & PSNR $\uparrow$ & SSIM $\uparrow$ & LPIPS $\downarrow$ & PSNR $\uparrow$ & SSIM $\uparrow$ & LPIPS $\downarrow$ & PSNR $\uparrow$ & SSIM $\uparrow$ & LPIPS $\downarrow$ & PSNR $\uparrow$ & SSIM $\uparrow$ \\
\midrule
DVGO       & 0.1371                     & 17.15                     & 0.9021                     & 0.1448                     & 17.23                     & 0.8945                     & 0.0793                     & 25.23                     & 0.9663                     & 0.1489                     & 15.70                     & 0.9050                     \\
3DGS       & 0.0490                     & 26.20                     & 0.9494                     & 0.0682                     & 16.16                     & 0.9388                     & 0.0374                     & 21.32                     & 0.9784                     & 0.1026                     & 15.19                     & 0.9331                     \\
DietNeRF   & 0.1264                     & 15.24                     & 0.9123                     & 0.0564                     & 16.74                     & 0.9472                     & 0.0253                     & 22.52                     & 0.9851                     & \cellcolor{metric-2}0.0373 & \cellcolor{metric-2}29.00 & \cellcolor{metric-2}0.9750 \\
RegNeRF    & 0.0772                     & 15.82                     & 0.9524                     & 0.1467                     & 14.15                     & 0.8969                     & 0.0293                     & 22.52                     & 0.9875                     & 0.1351                     & 15.80                     & 0.9186                     \\
\bottomrule
\end{tabular}
\end{table*}

\begin{table*}[thbp]
\newcommand{\xmark}{\ding{55}}
\centering
\caption{Comparisons of per-scene metrics of OmniObject3D with 9 input views.}
\label{tab: per scene omni3d 9.1}
\begin{tabular}{l|ccc|ccc|ccc}
\toprule
\multirow{2}{*}{Method} & \multicolumn{3}{c|}{backpack\_016} & \multicolumn{3}{c|}{box\_043} & \multicolumn{3}{c}{broccoli\_003} \\
& LPIPS $\downarrow$ & PSNR $\uparrow$ & SSIM $\uparrow$ & LPIPS $\downarrow$ & PSNR $\uparrow$ & SSIM $\uparrow$ & LPIPS $\downarrow$ & PSNR $\uparrow$ & SSIM $\uparrow$ \\
\midrule
DVGO       & 0.2299                     & 12.43                     & 0.8179                     & 0.0664                     & 22.20                     & 0.9677                     & 0.0683                     & 21.03                     & 0.9618                     \\
3DGS       & \cellcolor{metric-3}0.0519 & \cellcolor{metric-3}26.80 & \cellcolor{metric-3}0.9494 & 0.0506                     & 18.65                     & 0.9669                     & 0.0531                     & 16.45                     & 0.9552                     \\
DietNeRF   & 0.1806                     & 11.48                     & 0.8684                     & 0.0481                     & 18.68                     & 0.9656                     & 0.0995                     & 14.19                     & 0.9211                     \\
RegNeRF    & 0.2967                     & 10.46                     & 0.8276                     & 0.1051                     & 17.39                     & 0.9490                     & 0.0483                     & 16.84                     & 0.9663                     \\
FreeNeRF   & 0.2422                     & 12.39                     & 0.8084                     & 0.0530                     & 18.68                     & 0.9656                     & 0.0483                     & 16.84                     & 0.9625                     \\
SparseNeRF & 0.4932                     & 10.46                     & 0.7481                     & 0.0615                     & 18.59                     & 0.9639                     & 0.2476                     & 13.34                     & 0.8900                     \\
\bottomrule
\end{tabular}
\end{table*}

\begin{table*}[thbp]
\newcommand{\xmark}{\ding{55}}
\centering
\caption{Comparisons of per-scene metrics of OmniObject3D with 9 input views. \textit{continued}}
\label{tab: per scene omni3d 9.2}
\begin{tabular}{l|ccc|ccc|ccc|ccc}
\toprule
\multirow{2}{*}{Method} & \multicolumn{3}{c|}{picnic\_basket\_009} & \multicolumn{3}{c|}{pineapple\_013} & \multicolumn{3}{c|}{sandwich\_003} & \multicolumn{3}{c}{suitcase\_006} \\
& LPIPS $\downarrow$ & PSNR $\uparrow$ & SSIM $\uparrow$ & LPIPS $\downarrow$ & PSNR $\uparrow$ & SSIM $\uparrow$ & LPIPS $\downarrow$ & PSNR $\uparrow$ & SSIM $\uparrow$ & LPIPS $\downarrow$ & PSNR $\uparrow$ & SSIM $\uparrow$ \\
\midrule
DVGO       & 0.1501                     & 16.50                     & 0.9012                     & 0.0927                     & 18.63                     & 0.9393                     & 0.0509                     & 25.44                     & 0.9818                     & 0.0965                     & 17.72                     & 0.9430                     \\
3DGS       & \cellcolor{metric-2}0.0232 & \cellcolor{metric-3}32.98 & \cellcolor{metric-2}0.9768 & 0.0573                     & 16.73                     & 0.9483                     & 0.0316                     & 21.91                     & 0.9825                     & 0.1026                     & 15.19                     & 0.9331                     \\
DietNeRF   & 0.1416                     & 13.74                     & 0.8889                     & 0.1247                     & 14.14                     & 0.8914                     & 0.0253                     & 22.52                     & 0.9851                     & \cellcolor{metric-3}0.0325 & 30.99                     & 0.9799                     \\
RegNeRF    & 0.1746                     & 13.91                     & 0.8965                     & 0.0638                     & 16.74                     & 0.9534                     & 0.0293                     & 22.52                     & 0.9875                     & \cellcolor{metric-2}0.0317 & \cellcolor{metric-2}32.63 & \cellcolor{metric-1}0.9903 \\
\bottomrule
\end{tabular}
\end{table*}

\begin{table*}[thbp]
\newcommand{\xmark}{\ding{55}}
\centering
\caption{Comparisons of per-scene metrics of OpenIllumination with 4 input views. Methods with \textdagger means the metrics are from the ZeroRF paper~\cite{zerorf}.}
\label{tab: per scene oppo 4}
\begin{tabular}{l|ccc|ccc|ccc|ccc}
\toprule
\multirow{2}{*}{Method} & \multicolumn{3}{c|}{stone} & \multicolumn{3}{c|}{pumpkin} & \multicolumn{3}{c|}{toy} & \multicolumn{3}{c}{potato} \\
& LPIPS $\downarrow$ & PSNR $\uparrow$ & SSIM $\uparrow$ & LPIPS $\downarrow$ & PSNR $\uparrow$ & SSIM $\uparrow$ & LPIPS $\downarrow$ & PSNR $\uparrow$ & SSIM $\uparrow$ & LPIPS $\downarrow$ & PSNR $\uparrow$ & SSIM $\uparrow$ \\
\midrule
DVGO               & \cellcolor{metric-3}0.0829 & 21.99                     & 0.8957                     & 0.0945                     & 23.07                     & 0.9370                     & 0.0928                     & 21.69                     & 0.9121                     & 0.1196                     & 21.37                     & 0.9097                     \\
3DGS               & 0.2890                     & 10.93                     & 0.8107                     & 0.2950                     & 11.43                     & 0.8692                     & 0.2898                     & 11.16                     & 0.8213                     & 0.3063                     & 12.03                     & 0.8639                     \\
DietNeRF$^\dagger$ & 0.0850                     & \cellcolor{metric-2}24.05 & \cellcolor{metric-1}0.9210 & \cellcolor{metric-2}0.0600 & \cellcolor{metric-1}26.54 & \cellcolor{metric-1}0.9700 & \cellcolor{metric-2}0.0790 & \cellcolor{metric-1}24.98 & \cellcolor{metric-1}0.9490 & \cellcolor{metric-3}0.1030 & \cellcolor{metric-3}23.00 & \cellcolor{metric-1}0.9490 \\
RegNeRF$^\dagger$  & 0.4830                     & 10.26                     & 0.6020                     & 0.4650                     & 11.74                     & 0.7490                     & 0.4760                     & 10.04                     & 0.6370                     & 0.5050                     & 11.63                     & 0.7190                     \\
\bottomrule
\end{tabular}
\end{table*}

\begin{table*}[thbp]
\newcommand{\xmark}{\ding{55}}
\centering
\caption{Comparisons of per-scene metrics of OpenIllumination with 6 input views. Methods with \textdagger means the metrics are from the ZeroRF paper~\cite{zerorf}.}
\label{tab: per scene oppo 6}
\begin{tabular}{l|ccc|ccc|ccc|ccc}
\toprule
\multirow{2}{*}{Method} & \multicolumn{3}{c|}{stone} & \multicolumn{3}{c|}{pumpkin} & \multicolumn{3}{c|}{toy} & \multicolumn{3}{c}{potato} \\
& LPIPS $\downarrow$ & PSNR $\uparrow$ & SSIM $\uparrow$ & LPIPS $\downarrow$ & PSNR $\uparrow$ & SSIM $\uparrow$ & LPIPS $\downarrow$ & PSNR $\uparrow$ & SSIM $\uparrow$ & LPIPS $\downarrow$ & PSNR $\uparrow$ & SSIM $\uparrow$ \\
\midrule
DVGO               & \cellcolor{metric-2}0.0621 & 24.22                     & 0.9132                     & \cellcolor{metric-3}0.0682 & \cellcolor{metric-3}25.82 & 0.9527                     & \cellcolor{metric-3}0.0675 & 24.44                     & 0.9313                     & 0.0931                     & 24.24                     & 0.9279                     \\
3DGS               & 0.3139                     & 10.87                     & 0.7702                     & 0.2547                     & 12.98                     & 0.8823                     & 0.2993                     & 11.10                     & 0.7943                     & 0.3197                     & 11.70                     & 0.8462                     \\
DietNeRF$^\dagger$ & 0.0850                     & \cellcolor{metric-3}24.87 & \cellcolor{metric-3}0.9210 & 0.0730                     & 24.80                     & \cellcolor{metric-1}0.9660 & 0.0860                     & \cellcolor{metric-3}25.37 & \cellcolor{metric-3}0.9440 & \cellcolor{metric-3}0.0870 & \cellcolor{metric-3}25.63 & \cellcolor{metric-1}0.9550 \\
RegNeRF$^\dagger$  & 0.2880                     & 13.80                     & 0.8480                     & 0.3500                     & 13.58                     & 0.8480                     & 0.2370                     & 13.54                     & 0.8840                     & 0.3480                     & 13.92                     & 0.8540                     \\
\bottomrule
\end{tabular}
\end{table*}

\end{document}